\documentclass[runningheads]{llncs}
\usepackage[T1]{fontenc}
\usepackage{graphicx}
\usepackage{amsmath,amsfonts}
\usepackage{hyperref}
\usepackage{graphicx}
\usepackage{paralist}
\usepackage{array}
\usepackage{booktabs} 
\usepackage{multirow}
\usepackage{siunitx} 
\usepackage{textcomp}
\usepackage{stfloats}
\usepackage{url}
\usepackage{booktabs} 

\usepackage{color}    
\usepackage{verbatim}

\usepackage{cite}
\hypersetup{
    colorlinks=true,
    citecolor=blue
}
\usepackage{enumitem,kantlipsum}
\usepackage{subcaption}
\usepackage{tablefootnote}
\usepackage{cleveref}
\usepackage{svg}
\usepackage{textcomp}
\usepackage{breqn}
\usepackage{tikz}
\usepackage{adjustbox}

\usepackage{pgfplots}
\usepackage{booktabs}
\usepackage[linesnumbered,ruled,vlined]{algorithm2e}
\usepackage{algpseudocode}
\usepackage[utf8]{inputenc}
\usepackage{tikz}
\usepackage{amssymb}
\usepackage{pifont}

\usepackage{threeparttable}

\usepackage{wasysym}
\usepackage[table,xcdraw]{xcolor} 

\usepackage{orcidlink}
\usepackage{booktabs,makecell,threeparttable,tabularx,array,wasysym}
\newcolumntype{L}[1]{>{\raggedright\arraybackslash}p{#1}}
 \newcolumntype{C}[1]{>{\centering\arraybackslash}m{#1}}

\usepackage{colortbl}
\usepackage{xcolor}
\usepackage{subcaption}
\usepackage{booktabs}

\definecolor{dred}{RGB}{178,34,34}  
\definecolor{dgreen}{RGB}{34,110,34}

\usepackage{tikz}
\usetikzlibrary{
arrows.meta,
positioning,
shapes,
fit,
calc
}

\usepackage{tcolorbox}
\tcbuselibrary{breakable}

\definecolor{cNone}{RGB}{31,119,180}
\definecolor{cNormClip}{RGB}{174,199,232}
\definecolor{cKrum}{RGB}{255,127,14}
\definecolor{cTrim}{RGB}{255,187,120}
\definecolor{cScClip}{RGB}{44,160,44}
\definecolor{cWeakDP}{RGB}{152,223,138}
\definecolor{cAbalance}{RGB}{214,39,40}
\definecolor{cFLAME}{RGB}{255,152,150}
\definecolor{cDeepSight}{RGB}{148,103,189}
\definecolor{cBalance}{RGB}{197,176,213}
\definecolor{cSPP}{RGB}{140,86,75}
\definecolor{cMMAD}{RGB}{196,156,148}
\definecolor{cUBAR}{RGB}{227,119,194}
\definecolor{cDFLDual}{RGB}{247,182,210}

\newcommand{\none}{{\textcolor{cNone}{\ensuremath{\blacksquare}}~FedAvg}}
\newcommand{\normclip}{{\textcolor{cNormClip}{\ensuremath{\blacksquare}}~Norm-Clip}}
\newcommand{\krum}{{\textcolor{cKrum}{\ensuremath{\blacksquare}}~M-Krum}}
\newcommand{\trim}{{\textcolor{cTrim}{\ensuremath{\blacksquare}}~Trim}}
\newcommand{\scclip}{{\textcolor{cScClip}{\ensuremath{\blacksquare}}~SCCLIP}}
\newcommand{\weakdp}{{\textcolor{cWeakDP}{\ensuremath{\blacksquare}}~Weak-DP}}
\newcommand{\abalance}{{\textcolor{cAbalance}{\ensuremath{\blacksquare}}~ABALANCE}}
\newcommand{\flame}{{\textcolor{cFLAME}{\ensuremath{\blacksquare}}~FLAME}}
\newcommand{\deepsight}{{\textcolor{cDeepSight}{\ensuremath{\blacksquare}}~DeepSight}}
\newcommand{\balance}{{\textcolor{cBalance}{\ensuremath{\blacksquare}}~BALANCE}}
\newcommand{\spp}{{\textcolor{cSPP}{\ensuremath{\blacksquare}}~SPP}}
\newcommand{\mmad}{{\textcolor{cMMAD}{\ensuremath{\blacksquare}}~MMAD}}

\newcommand{\ubar}{{\textcolor{cUBAR}{\ensuremath{\blacksquare}}~UBAR}}

\newcommand{\dfldual}{{\textcolor{cDFLDual}{\ensuremath{\blacksquare}}~DFL-Dual}}

\newcommand{\ie}{\textit{i.e.}}
\newcommand{\eg}{\textit{e.g.}}

\usepackage{tikz}
\usetikzlibrary{
arrows.meta,
positioning,
shapes,
fit,
calc
}

\begin{document}
%

%
%
\title{BackDFL: A Unified Benchmark For Backdoor Attacks and Defenses In Decentralized Federated Learning}

\author{Mouhamed Amine Bouchiha\inst{1}\orcidID{0000-0001-6142-6855} \and Gregory Blanc\inst{1}\orcidID{0000-0001-8150-6617} \and
Yufei Han\inst{2}\orcidID{0000-0002-9035-6718}
}
\authorrunning{M. Bouchiha et al.}
%
\institute{SAMOVAR, Télécom SudParis, Institut Polytechnique de Paris \and PIRAT, INRIA Rennes, Rennes, France
}

\maketitle              
\begin{abstract}
Decentralized Federated Learning (DFL) promises trust-free collaborative learning by replacing the centralized parameter server with peer-to-peer model exchange. However, this architectural shift fundamentally reshapes the threat landscape. Without globally coordinated aggregation, DFL becomes particularly susceptible to backdoor attacks, in which malicious participants implant persistent hidden behaviors while maintaining high clean-task performance. In this paper, we argue that the robustness of DFL has been significantly overestimated. Existing studies rely on simplified threat models, non-adaptive adversaries, fragmented evaluation protocols, inconsistent communication topologies, and ad hoc training configurations, leading to an incomplete understanding of DFL security. To address these limitations, we present BackDFL, a unified benchmark for systematically evaluating DFL under realistic and adaptive backdoor attacks. Through extensive experiments, BackDFL exposes critical failure modes of decentralized learning. Our results demonstrate that both state-of-the-art Byzantine-robust DFL methods and adapted FL backdoor defenses fail under modest malicious participation rates (as low as 15\%), especially in heterogeneous settings, while their robustness varies substantially across communication graph topologies.

\keywords{Decentralized Federated Learning \and Backdoor Attacks \and Backdoor Defenses \and Unified Benchmark.}
\end{abstract}

\section{Introduction}
\label{sec:intro}

Federated Learning (FL)~\cite{mcmahan17} is a privacy-preserving distributed learning paradigm in which multiple participants collaboratively train a shared model without exchanging their raw data. Each client locally updates a copy of the current model using its private dataset and sends the resulting model update to a central server. The server then aggregates these updates to produce a new global model, which is redistributed to clients in the next training round. Despite its appeal, classical FL remains fundamentally centralized. A single server coordinates the training process, collects local model updates and performs aggregation to maintain the global model. This coordination introduces well-known drawbacks, including a single point of failure, performance bottlenecks, and strong trust dependencies, which can expose the system to attacks and disruptions.
To address these limitations, recent work has shifted toward Decentralized Federated Learning (DFL)~\cite{beltran2023decentralized,lalitha2018fully, el2021collaborative, guo2021byzantine, liu2023prometheus,kalra2023decentralized, hegedHus2019gossip}, where model aggregation is performed over peer-to-peer (P2P) or partially structured networks without relying on a central coordinator. By eliminating the single point of failure and centralized trust, DFL provides improved fault tolerance and better matches the requirements of large-scale, dynamic environments such as vehicular networks and IoT ecosystems~\cite{yuan2024decentralized}. This paradigm has demonstrated promise across a wide range of applications. In autonomous driving, for example, connected vehicles~\cite{chen2021bdfl, bouchiha2026automated, chellapandi2023federated} can collaboratively learn perception models from locally captured images or videos to detect traffic signs, lanes, pedestrians, and other critical objects. However, decentralization also removes global oversight mechanisms that traditionally support robust aggregation and coordinated attack detection. As a result, under weaker trust assumptions, learning becomes vulnerable to adaptive and stealthy poisoning.

In this context, adaptive backdoor attacks (BAs) become particularly insidious~\cite{A3FL, IBA, F3BA, neurotoxin}. By crafting malicious local updates that survive decentralized aggregation, adversaries can implant hidden behaviors into each peer's global model—preserving high accuracy on benign data while inducing strong targeted misclassification on trigger inputs. Although several Byzantine-robust DFL methods have been proposed~\cite{guo2021byzantine, he2022byzantine, fan2025bad, sun2024byzantine}, most evaluate their defenses using naive backdoor scenarios~\cite{BadNets}, \cite{ModelReplacement} that fail to capture realistic, adaptive threats. In particular, these evaluations assume static triggers~\cite{fang2024byzantine}, stationary adversaries, and often ignore the underlying network topology~\cite{sun2024byzantine}, which significantly underestimates the potential of sophisticated attackers (\eg, A3FL~\cite{A3FL}, IBA~\cite{IBA}) who can craft updates that survive aggregation. As a result, the reported robustness of existing methods may be overly optimistic, leaving critical vulnerabilities unaddressed. Moreover, the broader DFL literature suffers from methodological inconsistencies that further complicate meaningful comparison. Studies employ heterogeneous communication topologies (ring, regular, small-world,...)~\cite{fang2024byzantine}, \cite{he2022byzantine} and widely varying local training hyperparameters (epochs, learning rates, optimizers, batch sizes)~\cite{sun2024byzantine}, ~\cite{guo2021byzantine}, even though both attack propagation and defense effectiveness are highly sensitive to these factors. Compounding these issues, most DFL defenses lack publicly available and reproducible implementations (\eg, BALANCE~\cite{fang2024byzantine}, UBAR~\cite{guo2021byzantine}, DFL-Dual~\cite{sun2024byzantine}), with only a few exceptions such as LEARN~\cite{el2021collaborative} and SCCLIP~\cite{he2022byzantine}.

In contrast to this fragmented state of DFL research, backdoor defense in centralized FL is far more mature and systematically studied. Numerous defense mechanisms have been proposed for centralized FL (\eg, FLAME~\cite{Flame}, DeepSight~\cite{rieger2022deepsight}, MMAD~\cite{huang2023multi}), benefitting from a parameter server that has global visibility over all selected client updates. However, such server-assisted defenses cannot be directly applied in fully DFL systems~\cite{fang2024byzantine}, where no participant has a global view. This could make centralized strategies either inapplicable or highly inefficient when executed locally. Thus, their effectiveness needs to be systematically evaluated when adapted to the DFL setting. These fundamental inconsistencies and gaps motivate a deeper investigation into the security limits of DFL under BAs. The main \textbf{contributions} of this paper are as follows:
\begin{enumerate}
 \item We present a unified and reproducible benchmark\footnote{\url{https://github.com/mohaminemed/BackDFL}} for backdoor robustness in fully decentralized FL.
 \item We conduct an extensive comparative evaluation of thirteen defenses showing their failure modes.
\end{enumerate}
The remainder of this paper is organized as follows. Section~\ref{sec:problemdef} introduces the problem formulation and outlines the key motivations. Section~\ref{sec:related_work} presents the related work. Section~\ref{sec:dfl-system-threat} defines the DFL system and threat models. Section~\ref{sec:BackDFL} presents the proposed BackDFL, a unified benchmark to assess the vulnerability of DFL to BAs. Section~\ref{sec:experiments} presents the evaluation methodology and discusses the results. Section~\ref{sec:conclusion} concludes the paper.

\section{Related Work} \label{sec:related_work}

Backdoor attacks in FL have been extensively studied in the \emph{centralized} setting~\cite{nguyen2024backdoor, dao2025backfed, bellachia2026sok}, where a parameter server aggregates client updates. For instance, Nguyen et al.~\cite{nguyen2024backdoor} offer focused overviews of attack strategies and defenses. More recent efforts, such as BackFed~\cite{dao2025backfed}, provide large-scale benchmarking frameworks that highlight implementation pitfalls and evaluation inconsistencies. Despite these advances, existing works remain fundamentally tied to the assumptions of centralized FL (CFL), where a trusted server orchestrates aggregation and often supports defense mechanisms (\eg, server-side validation, robust aggregation rules). However, \textit{BAs in DFL remain largely underexplored from an evaluation perspective}. This gap is critical as insights derived from centralized FL cannot be directly extrapolated to DFL due to several fundamental differences: (i) \emph{decentralized aggregation}, where each client performs local aggregation using its own reference model; (ii) \emph{topology-constrained communication}, which controls how poisoned updates propagate through the network; and (iii) \emph{increased attack surface}, where adversaries can exploit graph structure. These factors can amplify or suppress BA persistence in ways that are not yet studied.

Beyond methodological gaps, prior studies employ heterogeneous experimental configurations that hinder reproducibility and comparability. Works differ substantially in graph topology, local training schedules, optimizers, and hyperparameters~\cite{fang2024byzantine, he2022byzantine, sun2024byzantine, guo2021byzantine}. Since both benign convergence and malicious diffusion are highly sensitive to these factors, reported robustness claims cannot be meaningfully contrasted across papers. Threat models are also simplified. Most evaluations rely on static triggers, fixed poisoned datasets, and non-adaptive adversaries~\cite{fang2024byzantine, sun2024byzantine}, thereby underestimating the capabilities of realistic attackers in decentralized settings.

\section{Problem Definition}
\label{sec:problemdef}

The transition to fully DFL addresses the limitations of conventional centralized FL by replacing the parameter server with direct P2P model exchange over a communication graph for enhanced resilience and autonomy. However, this design shift does not address persistent security challenges, particularly \emph{backdoor attacks}, which remain largely unexplored. 

\textbf{Research Questions.} There is a clear need for a rigorous, reproducible, and scientifically sound benchmark for evaluating BAs in DFL, which should \textit{(1) unify DFL configurations and training protocols, (2) harmonize attack definitions and threat models, (3) address implementation pitfalls and ensure consistent evaluation metrics, and (4) enable fair, comparable, and reproducible assessment of both attacks and defenses}. Guided by this scientific requirement, this work proposes BackDFL to address the following questions:
\begin{itemize}
\item \textbf{RQ1.} Which centralized FL defenses remain effective when adapted to DFL, and why do others fail?
\item \textbf{RQ2.} How do established stealthy BAs (\eg, \texttt{Neurotoxin}, \texttt{A3FL}, \texttt{IBA}) behave in DFL settings?
\item \textbf{RQ3.} What is the impact of data and model characteristics on robustness?
\item \textbf{RQ4.} What is the impact of the DFL graph topology?
\end{itemize}

\begin{figure}[t]
    \centering
    \includegraphics[width=0.85\linewidth]{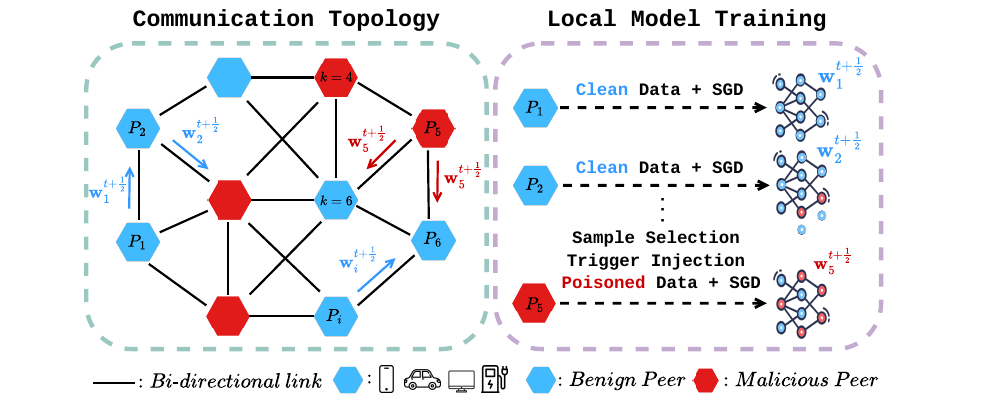}
    \caption{Backdoor propagation in DFL through P2P model exchange.}
    \label{fig:badfl}
\end{figure}

\section{System \& Threat Models}
\label{sec:dfl-system-threat}

This section defines the DFL system and threat models we adopt throughout this paper.

\subsection{System Model}
We consider a DFL setting following prior work~\cite{fang2024byzantine, sun2024byzantine, pasquini2023security}, where a set of clients 
$V$ is connected through an undirected communication graph $G=(V,E)$, with $(i,j)\in E$ indicating that clients $i$ and $j$ 
can directly exchange model parameters as depicted in Figure~\ref{fig:badfl}. Each client $i\in V$ holds a private local dataset $\mathcal{D}_i$ and jointly aims to learn a global model $\mathbf{w} \in \theta \subset \mathbb{R}^d$ that minimizes the empirical risk:
\begin{equation}
\label{eq:erm}
\mathbf{w}^\ast = \arg\min_{\mathbf{w} \in \theta} 
F(\mathbf{w}) = \frac{1}{|\mathcal{D}|} \sum_{\zeta \in \mathcal{D}} f(\mathbf{w}, \zeta),
\end{equation}
where $\mathcal{D} = \biguplus_{i \in V} \mathcal{D}_i$.

Unlike classical federated learning, DFL operates \emph{without a central server}. Each client maintains a local model 
$\mathbf{w}_i$ and interacts only with its neighbors $\mathcal{N}_i = \{j \mid (i,j)\in E\}$. At round $t$, client $i$ first performs local optimization  (\eg, SGD) using a learning rate $\eta$ to obtain
\begin{equation}
\label{eq:dfl-train}
\mathbf{w}_{i,t+\frac{1}{2}} = \textsc{LocalTraining}(\mathbf{w}_{i,t}, \mathcal{D}_i, \eta),
\end{equation}
then aggregates neighbors’ models via
\begin{equation}
\label{eq:dfl-update}
\mathbf{w}_{i,t+1} 
= \alpha \mathbf{w}_{i,t+\frac{1}{2}}
+ (1-\alpha)\,\textsc{AGG}\left(\{\mathbf{w}_{j,t+\frac{1}{2}} \mid j \in \mathcal{N}_i\}\right),
\end{equation}
where $\alpha \in [0,1]$ controls the trade-off between self-reliance and consensus, and $\textsc{AGG}(\cdot)$ denotes the aggregation rule (\eg, FedAvg~\cite{mcmahan17}, Median~\cite{yin2018byzantine}). Overall, each round consists of 
\begin{inparaenum}[(i)]
\item \emph{local training and model exchange} using Eq.\eqref{eq:dfl-train}, followed by
\item \emph{decentralized aggregation} using Eq.\eqref{eq:dfl-update}.
\end{inparaenum}

\begin{figure}[t]
    \centering
    \includegraphics[width=0.9\linewidth]{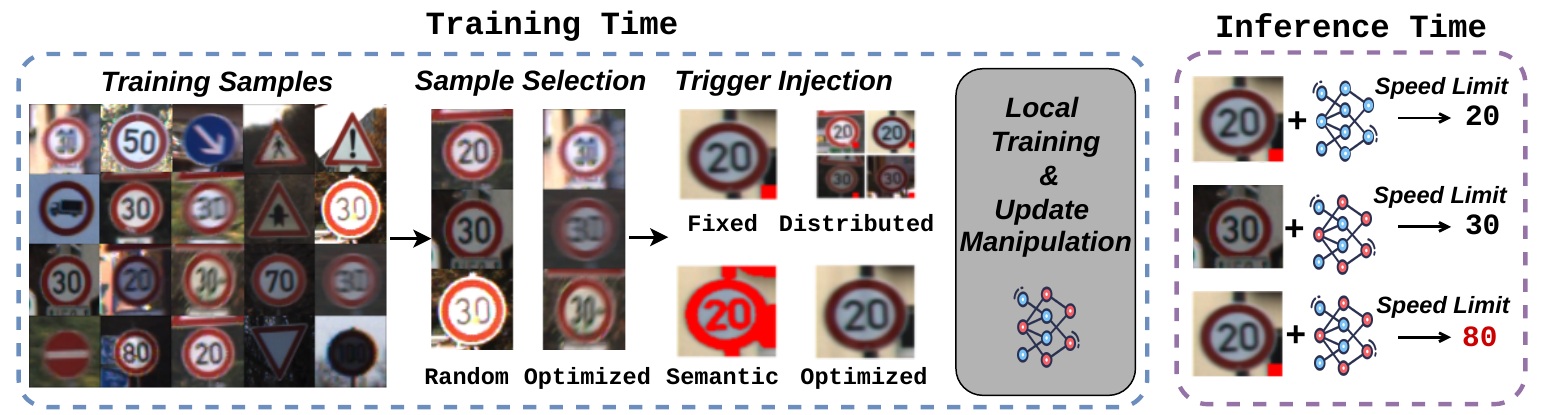}
    \caption{Overview of the backdoor attack workflow on the \texttt{GTSRB} dataset, the \texttt{training-time} stage (\textbf{left}) (sample selection, trigger injection, and local training on poisoned 
data) and the \texttt{inference-time} stage (\textbf{right}), where the trigger is inserted to activate the attack.}
\label{fig:BApipline}
\end{figure}

\subsection{Threat Model}

Following prior work on DFL security~\cite{fang2020local, guo2021byzantine, he2022byzantine, pasquini2023security, fang2024byzantine}, 
we consider an \emph{active adversary} controlling a subset of clients in the decentralized network.

\noindent \textbf{Attacker Objective.} The attacker aims to poison the training process to implant \emph{backdoors} (as shown in Figure~\ref{fig:BApipline}) into model updates propagated to benign clients, while remaining stealthy.

\noindent \textbf{Attacker Capabilities.} We consider an adversarial scenario in which the attacker can compromise a fraction $\beta_M$ (default 15\%) of the client population. Malicious clients can \begin{inparaenum}[(i)]
\item inject poisoned samples into their local datasets, and/or
\item manipulate their model updates before sharing them with neighbors.
\end{inparaenum}
The communication topology $G$ is fixed and cannot be modified by the attacker, although benign clients may temporarily disconnect due to network instability.

\noindent \textbf{Attacker and Defender Knowledge.} We assume an \emph{adaptive attacker} with full knowledge of \begin{inparaenum}[(i)]
\item the compromised clients’ training data,
\item their aggregation rules and hyperparameters.
\end{inparaenum} The attacker has no prior knowledge of the communication graph $G$ or the neighborhood structure of individual clients, and can only operate based on locally observed information. Both benign and malicious clients can naturally observe their neighbors’ shared model updates. In contrast, the \emph{defender} does not know which clients are malicious, has no visibility into the attack strategy, and lacks knowledge of both global and local malicious ratios.

\section{Proposed BackDFL}
\label{sec:BackDFL}

We present a modular framework for evaluating BAs and BDs in centralized and decentralized FL. Implemented in Python/PyTorch, it emphasizes reproducibility and configurability, enabling easy integration of new attacks, defenses, and datasets. BackDFL’s design goals are:

\begin{itemize}
    \item \textbf{Modularity:} Each functional component---datasets, models, attacks, defenses, and experiment flows---is encapsulated in independent modules to enable seamless extension and integration.
    \item \textbf{Reproducibility:} The framework is built to replicate the results of this study while providing a stable base for future research.
    \item \textbf{FL and DFL support:} In addition to centralized FL, it supports fully decentralized FL with parallel training and configurable graphs.
     \item \textbf{Configuration-driven experiments:} Experiments are defined via YAML configuration files to eliminate the need for repetitive boilerplate code.
    \item \textbf{Rich algorithmic implementations:} Includes well established BAs (\texttt{DBA}, \texttt{Scaling}, \texttt{Neurotoxin}, \texttt{A3FL}, \texttt{IBA}) and defenses (BALANCE, SCCLIP, M-Krum, Weak-DP, FLAME, DeepSight,...), several benchmark datasets (\texttt{MNIST}, \texttt{CIFAR-10/100}, \texttt{GTSRB}, \texttt{FEMNIST}, \texttt{HAR}, ...) along with their standard models.
\end{itemize}

\begin{figure}[t]
    \centering
    \includegraphics[width=0.8\linewidth]{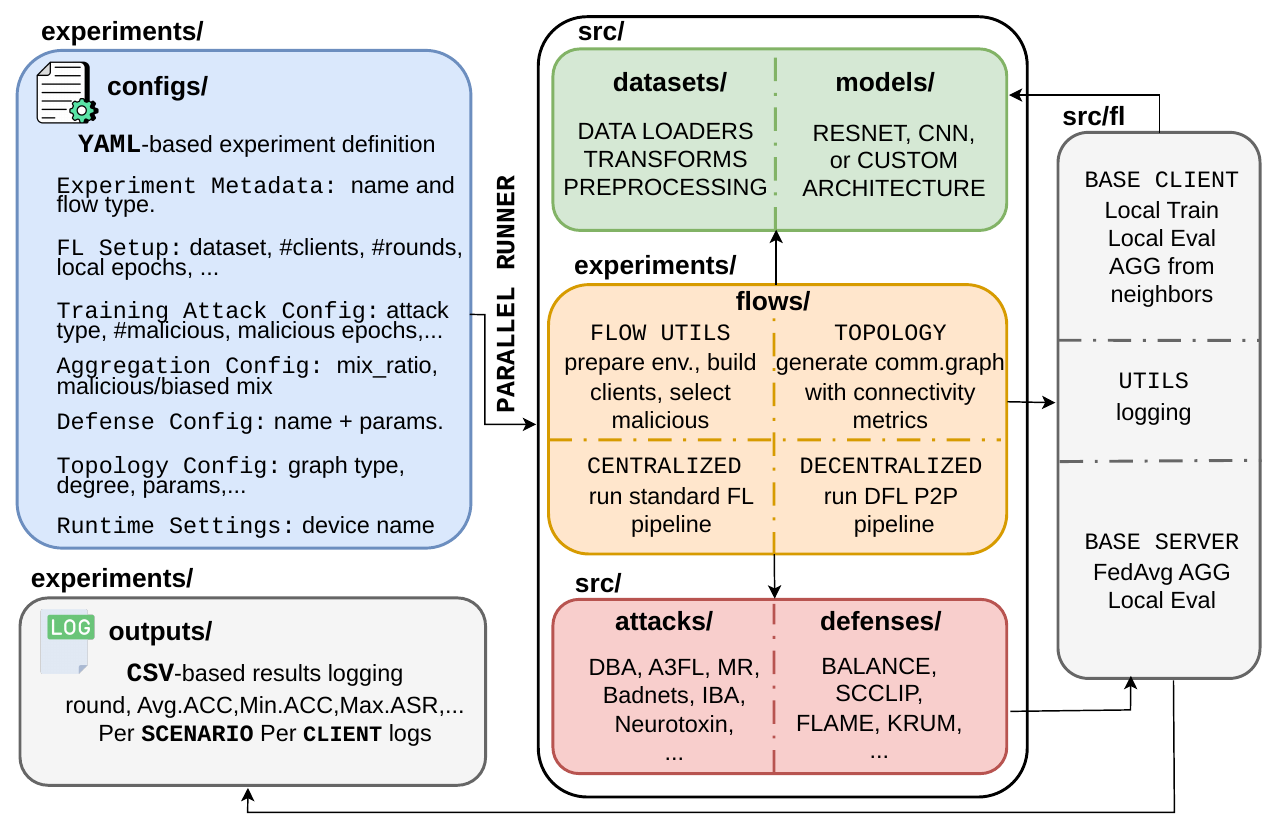}
        \caption{BackDFL: A framework for BAs and BDs in DFL. The architecture is modular, with configuration files feeding experiment flows, which interact with datasets, models, attack and defense modules. Utility modules are shared.}
    \label{fig:BackDFL}
\end{figure}

\subsubsection{Framework Overview.} The framework follows a modular design (Figure~\ref{fig:BackDFL}) comprising \begin{inparaenum}[(i)]
\item a \textit{configuration layer} with YAML-based experiment definitions feeding both CFL and DFL flows,
\item a \textit{data \& model layer} handling datasets and model architectures,
\item an \textit{experiment flow layer} with two parallel pipelines: \texttt{centralized} flow for CFL and \texttt{decentralized} flow for DFL with topology management,
\item an \textit{attack \& defense layer} implementing BAs and defense mechanisms shared across both flows,
\item a \textit{base FL layer} providing foundational client and server code, and
\item an \textit{output layer} computing performance metrics while offering utility logs for iterative experiments.
\end{inparaenum} BackDFL design encourages modular extension: new attacks, defenses, datasets and models can be plugged in without modifying core experiment flows.

\subsubsection{Implementation Scope.} At the time of writing, BackDFL integrates \textbf{six} well-established targeted BAs (fixed trigger~\cite{BadNets, DBA}, optimized trigger~\cite{A3FL, IBA}, gradient manipulation~\cite{neurotoxin, ModelReplacement}), \textbf{five} classical untargeted attacks, and \textbf{fourteen} defenses (9 adapted FL BDs and 5 DFL-specific robust methods). BackDFL supports evaluation on \textbf{seven} datasets. Although the key focus of this work is the vulnerability of DFL to adaptive backdoors, we consider it important to include classical untargeted attacks as well, since Byzantine robustness encompasses both. These include Label Flipping~\cite{Virat2021}, Feature~\cite{wang2021feature}, Gauss~\cite{blanchard2017machine}, Krum and Trim~\cite{fang2020local} attacks.

\section{Evaluation \& Results}
\label{sec:experiments}
This section presents the empirical evaluation of several FL BDs and Byzantine-robust DFL methods under several BAs using the proposed BackDFL. 

\subsection{Experimental Setup}

\noindent\textbf{Datasets and models.}
We use BackDFL to evaluate BAs and BDs on seven benchmark datasets: \texttt{MNIST}~\cite{lecun1998mnist}, \texttt{FashionMNIST}~\cite{xiao2017fashion}, \texttt{FEMNIST}~\cite{emnist}, \texttt{GTSRB}~\cite{gtsrb}, \texttt{CIFAR-10} and \texttt{CIFAR-100}~\cite{cifar}, as well as \texttt{HAR}~\cite{anguita2013public}, using adapted neural network architectures under non-IID settings. 

For \texttt{MNIST}, we use a SimpleCNN model. For \texttt{FashionMNIST} (28$\times$28 grayscale images), we employ a lightweight convolutional neural network (Fashion-CNN) composed of two convolutional blocks followed by two fully connected layers. For the \texttt{FEMNIST} dataset (28$\times$28 grayscale images), we employ a LeNet-5 model. For \texttt{CIFAR-10} and \texttt{CIFAR-100} (32$\times$32 color images), we use a ResNet18 model, and for \texttt{GTSRB} (color traffic sign images), we use a custom CNN. For \texttt{HAR}, which consists of 561-dimensional feature vectors extracted from smartphone sensors, we adopt a multilayer perceptron (MLP) with two hidden layers (256 and 128 units). Appendix~\ref{app:hyper} provides the complete training configurations.\\
\noindent\textbf{Attacks.}
We evaluate six \textit{well-established} BAs selected to cover a diverse spectrum of strategies, including input-space triggers, learned triggers, gradient manipulation, and scaling-based model replacement. This diversity ensures that our benchmark captures both classical~\cite{BadNets, ModelReplacement, DBA} and adaptive attack behaviors~\cite{neurotoxin, A3FL, IBA}.
For all attacks, the poisoning rate is 25\% of local data (as in \texttt{A3FL}~\cite{A3FL}), the attacker fraction $\beta_M$ is set to 15\% of the population, and all poisoned samples are relabeled to the target class prior to training.\\
\noindent\textbf{Defenses}. We evaluate both CFL and DFL defenses against these attacks. The selection of FL defense methods is aligned with the DFL system model presented in Section~\ref{sec:dfl-system-threat}. For CFL defenses, we focus on \textbf{pre-}~\cite{Flame, rieger2022deepsight, huang2023multi, wang2025can}, \textbf{in-}~\cite{blanchard2017machine, yin2018byzantine, sun2019can}, and \textbf{post-aggregation}~\cite{sun2019can, Flame}  methods that are compatible with DFL, \ie, they can be instantiated locally by each client acting as a \textit{defensive aggregator}. Methods that rely on global signals (a root dataset~\cite{cao2020fltrust}) or post-hoc model purification~\cite{wu2021adversarial} are thus excluded, as their assumptions do not hold in DFL settings. This ensures that the evaluation measures the effectiveness of defenses that can realistically be deployed in DFL. For DFL-specific defenses, we cover all three design paradigms: peer-based trust assessment~\cite{sun2024byzantine}, filtering via distance- or loss-based acceptance tests~\cite{fang2024byzantine, guo2021byzantine}, and norm-based update bounding~\cite{he2022byzantine}. We exclude defenses such as BaDFL~\cite{yuan2025badfl} and LEARN~\cite{el2021collaborative} from the evaluation since they directly modify the local training procedure or client optimization dynamics and therefore do not conform to our system model. The final selected set includes: \textbf{DFL methods} ---\ubar{}~\cite{guo2021byzantine}, \scclip{}~\cite{he2022byzantine}, \balance{}~\cite{fang2024byzantine}, \dfldual{}~\cite{sun2024byzantine}, \abalance{}; \textbf{CFL defenses}--- \deepsight{}~\cite{rieger2022deepsight}, \flame{}~\cite{Flame}, \spp{}~\cite{wang2025can}, \mmad{}~\cite{huang2023multi}, \normclip{}, \weakdp{}~\cite{sun2019can}, \krum{}~\cite{blanchard2017machine}, and \trim~\cite{yin2018byzantine}. Detailed descriptions are available in our public repository. \abalance{} extends \balance{} by replacing its acceptance rule with an \textit{adaptive} threshold that accounts for both the distribution of received updates and their temporal evolution:
\[
\tau_{\text{adaptive}}^{(t)}
=
\min\!\left(
\mathrm{median}(d_{ij}^{(t)})+\sigma,\,
\tau_{\text{prev}}
\right)
\cdot
\frac{1}{1+\gamma\sigma\lambda(t)},
\]
where $\sigma$ denotes the median absolute deviation (MAD) of the received distances and $\tau_{\text{prev}}$ is the threshold from the previous round. Compared with the original BALANCE, this adaptive criterion relaxes acceptance during early rounds under highly non-IID data while progressively tightening the threshold through temporal consistency to improve robustness against stealthy BAs.\\

\noindent\textbf{Metrics.} Following prior work on FL and DFL security, we report both utility and attack effectiveness metrics:
\begin{itemize}
\item \textbf{Main-task accuracy (ACC):} the average (Avg.ACC) and (Min.ACC) accuracy on clean test data across all benign clients~\cite{A3FL, IBA}.
\item \textbf{Attack success rate (ASR):} the maximum ASR (Max.ASR) observed among all benign clients, defined as the fraction of triggered inputs (\ie, poisoned test samples) misclassified to the target label~\cite{fang2024byzantine, A3FL}.
\item \textbf{Durability:} we measure the maximal ASR achieved at the end of training (Final.ASR), as well as the backdoor Lifespan, defined as the number of communication rounds required for Max.ASR to drop below a predefined threshold (\eg, $0.5$)~\cite{neurotoxin, A3FL}. In the default setting, the attack is activated for a fixed window ($[t_{\text{start}}, t_{\text{end}}]$) in order to report both metrics consistently.
\end{itemize}

\noindent \textbf{Default Settings.} All experiments are conducted in a DFL setup, where clients exchange model updates exclusively with their neighbors according to a P2P communication topology (Section~\ref{sec:dfl-system-threat}). Following prior work on robust DFL~\cite{fang2024byzantine,pasquini2023security}, we consider a regular communication graph with uniform degree, stable connectivity, and controlled spectral properties (default \texttt{regular-(20-10)}). Here, $|V| = N = 20,\quad \deg(v) = k,\ \forall v \in V$, with all clients participating in every communication round. We evaluate two local training regimes: (i) a \emph{single-epoch} mode with one epoch per round (communication-intensive), and (ii) a \emph{few-epoch} mode with 5 local epochs per round (communication-efficient).

Data heterogeneity is introduced via a Dirichlet partitioning with concentration parameter $Dirichlet$-$\alpha = 0.5$, a standard choice for DFL. Clients train with mini-batches of size 32, while model evaluation uses a batch size of 256. Unless otherwise stated, all experiments adopt a learning rate of 0.01 and, as in Fang et al.~\cite{fang2024byzantine}, a mixing ratio $\alpha$ of 0.5 when aggregating incoming neighbor updates. By default, we assume that all benign clients use the same $\alpha$, but unlike Fang et al.~\cite{fang2024byzantine}, we assume different initial models for different clients, as this is the most realistic setup. Finally, we use a uniform environment (two identical NVIDIA L40S GPUs) to ensure consistent computational conditions. The above configurations represent the default settings; any deliberate changes (\eg, attack mode) are indicated in the corresponding sections.  

\subsection{Results and Analysis}
We organize our evaluation in accordance with the \textbf{RQs} introduced in Section~\ref{sec:problemdef}. Sensitivity analyses with respect to key experimental factors including malicious rate, data heterogeneity, and mixing ratio are provided in Appendix~\ref{app:factors}.

\begin{figure}[t]
    \centering
    \subfloat[CFL; rand-($N,k$) denotes random subsampling of k clients.]{
        \includegraphics[width=0.75\columnwidth]{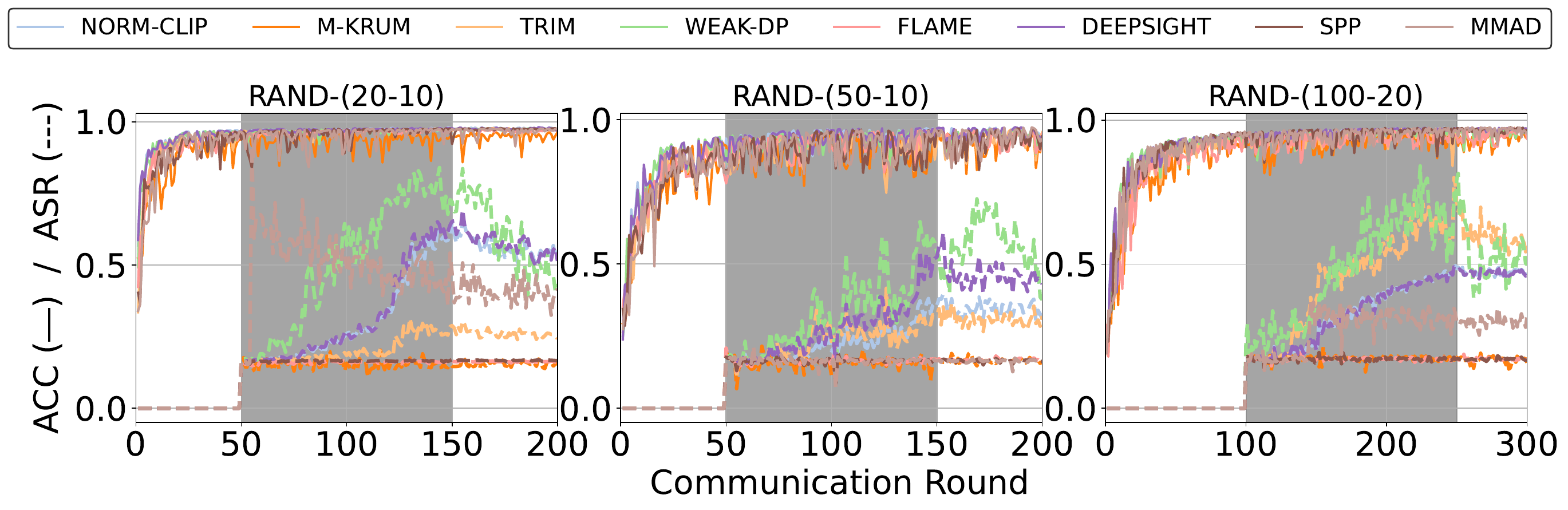}
        \label{fig:cfl_settings}
    }
    \hfill
    
    \subfloat[DFL; regular-($N,k$) denotes a random k-regular graph.]{
        \includegraphics[width=0.75\columnwidth]{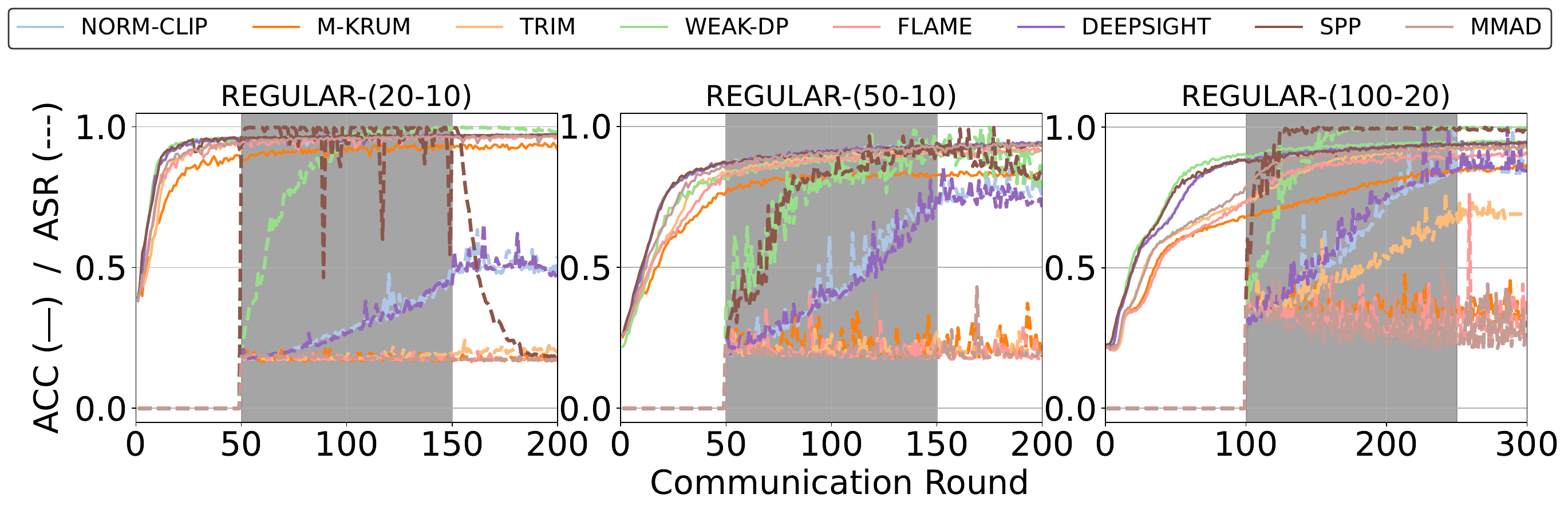}
        \label{fig:dfl_settings}
    }
    \caption{Impact of the FL setting (CFL vs. DFL) on robustness with identical attack settings (\texttt{Neurotoxin} on \texttt{HAR}, 15\% malicious).}
    \label{fig:flsettings}
\end{figure}

\noindent\textbf{\textbf{RQ1.} Which centralized FL defenses remain effective when adapted to DFL, and why do others fail?}
Figures~\ref{fig:cfl_settings} and \ref{fig:dfl_settings} highlight a clear discrepancy between centralized and decentralized FL under identical attack settings.

\textit{Centralized FL.}
All methods maintain high accuracy ($\geq 0.94$), but robustness differs significantly. \krum{}, \flame{}, and \spp{} effectively suppress the attack, achieving low ASR ($\approx 0.16$--$0.21$) and zero durability, indicating complete removal of malicious influence. In contrast, \normclip{}, \weakdp{}, and \deepsight{} exhibit persistent vulnerabilities, with moderate to high ASR (up to $0.84$) and maximum durability, despite preserving accuracy. Notably, performance remains stable across network scales with a stable effect of aggregation.

\textit{Decentralized FL.}
Robustness degrades substantially in all configurations. While average accuracy remains acceptable, most defenses fail to control attack propagation. Even strong centralized defenses (\krum{}, \flame{}, \mmad{}) show non-negligible ASR and increased durability, especially in larger networks. More critically, \weakdp{}, \deepsight{}, and \spp{} collapse entirely, reaching near-perfect ASR ($\approx 1.0$) and maximal durability with uncontrolled diffusion of malicious updates.  This degradation arises from the fundamentally local nature of DFL: \begin{inparaenum}[(i)]
    \item each client observes only a limited set of neighbors, preventing reliable statistical detection of anomalous updates and
    \item as malicious updates propagate, the local reference model itself becomes progressively contaminated, causing adversarial updates to appear consistent with the neighborhood.
\end{inparaenum}
\begin{tcolorbox}[breakable, boxrule=0pt,
  colframe=gray!40]
\textbf{Takeaways.} 
\textit{Decentralization fundamentally alters the threat dynamics}: local aggregation errors accumulate and propagate across neighbors, enabling persistent backdoor injection. Consequently, defenses effective in centralized FL do not directly transfer to decentralized settings.
\end{tcolorbox}

\begingroup
\renewcommand{\arraystretch}{1.05}
\setlength{\tabcolsep}{3.5pt}
\begin{table*}[t]
\centering
\caption{Per-attack, per-defense results on \texttt{CIFAR-10} and \texttt{GTSRB} under IID and
non-IID client partitions. Each cell reports \texttt{Min.ACC/Max.ASR}.
Lowest \texttt{ASR} per attack--dataset--partition block is \textbf{bold};
worst \texttt{Min.ACC} per attack--dataset--partition block is
\underline{underlined}. Failure cases with
\texttt{Max.ASR} $\geq 0.500$ are \textcolor{red}{red} shaded.}
\label{tab:rq2-results}

\resizebox{0.78\textwidth}{!}{%
\begin{tabular}{ll cc cc cc cc}
\toprule
& &
\multicolumn{4}{c}{\texttt{CIFAR-10}} &
\multicolumn{4}{c}{\texttt{GTSRB}} \\
\cmidrule(lr){3-6}
\cmidrule(lr){7-10}
& &
\multicolumn{2}{c}{\textbf{IID}} &
\multicolumn{2}{c}{\textbf{Non-IID}} &
\multicolumn{2}{c}{\textbf{IID}} &
\multicolumn{2}{c}{\textbf{Non-IID}} \\
\cmidrule(lr){3-4}
\cmidrule(lr){5-6}
\cmidrule(lr){7-8}
\cmidrule(lr){9-10}
\textbf{Attack} & \textbf{Defense}
& \texttt{Min.ACC} & \texttt{Max.ASR}
& \texttt{Min.ACC} & \texttt{Max.ASR}
& \texttt{Min.ACC} & \texttt{Max.ASR}
& \texttt{Min.ACC} & \texttt{Max.ASR} \\
\midrule

\multirow{13}{*}{\rotatebox{90}{\Large \texttt{Neurotoxin}}}

 & \normclip{}
 & 0.847 & 0.245
 & 0.789 & 0.258
 & 0.944 & 0.357
 & 0.937 & 0.291 \\
 
 & \krum{}
 & 0.841 & 0.128
 & 0.745 & 0.205
 & \underline{0.928} & 0.060
 & 0.899 & \cellcolor{red!15}0.773 \\

 & \trim{}
 & 0.845 & 0.138
 & 0.792 & 0.261
 & 0.943 & 0.110
 & 0.941 & 0.115 \\

 & \scclip{}
 & 0.840 & \cellcolor{red!15}0.661
 & 0.748 & \cellcolor{red!15}0.694
 & 0.941 & \cellcolor{red!15}0.627
 & 0.906 & \cellcolor{red!15}0.832 \\

 & \weakdp{}
 & 0.842 & 0.254
 & 0.778 & 0.269
 & 0.944 & \cellcolor{red!15}0.566
 & 0.940 & \cellcolor{red!15}0.531 \\

 & \abalance{}
 & 0.837 & 0.196
 & 0.760 & 0.485
 & 0.941 & 0.045
 & 0.929 & 0.095 \\

 & \flame{}
 & 0.845 & 0.148
 & 0.775 & 0.239
 & 0.932 & 0.106
 & 0.930 & \textbf{0.064} \\

 & \deepsight{}
 & \underline{0.810} & 0.155
 & 0.757 & 0.223
 & 0.934 & 0.295
 & 0.884 & 0.277 \\

 & \balance{}
 & 0.843 & \cellcolor{red!15}0.512
 & 0.766 & \cellcolor{red!15}0.553
 & 0.937 & 0.079
 & 0.894 & \cellcolor{red!15}0.511 \\

 & \spp{}
 & 0.832 & 0.129
 & 0.757 & 0.426
 & 0.942 & \textbf{0.043}
 & 0.939 & \textbf{0.064} \\

 & \mmad{}
 & 0.852 & 0.142
 & 0.790 & \textbf{0.178}
 & 0.934 & 0.286
 & 0.912 & \cellcolor{red!15}0.810 \\

 & \ubar{}
 & 0.823 & \textbf{0.128}
 & \underline{0.719} & 0.226
 & 0.929 & 0.053
 & \underline{0.795} & 0.073 \\

 & \dfldual{}
 & 0.839 & 0.131
 & 0.781 & 0.213
 & 0.934 & 0.055
 & 0.912 & 0.456 \\

\midrule

\multirow{13}{*}{\rotatebox{90}{\Large \texttt{A3FL}}}

 & \normclip{}
 & 0.846 & 0.235
 & 0.775 & 0.388
 & 0.940 & \cellcolor{red!15}0.582
 & 0.936 & \cellcolor{red!15}0.527 \\
 
 & \krum{}
 & 0.832 & 0.370
 & 0.730 & \cellcolor{red!15}0.776
 & 0.941 & 0.049
 & 0.925 & \cellcolor{red!15}0.851 \\

 & \trim{}
 & 0.846 & 0.256
 & 0.792 & 0.417
 & 0.944 & 0.420
 & 0.938 & 0.390 \\

 & \scclip{}
 & 0.841 & 0.371
 & 0.752 & 0.388
 & 0.941 & \cellcolor{red!15}0.672
 & 0.935 & \cellcolor{red!15}0.898 \\

 & \weakdp{}
 & 0.843 & 0.224
 & 0.781 & 0.408
 & 0.941 & \cellcolor{red!15}0.650
 & 0.935 & \cellcolor{red!15}0.605 \\

 & \abalance{}
 & 0.837 & 0.464
 & 0.768 & 0.486
 & 0.942 & 0.048
 & 0.936 & 0.074 \\

 & \flame{}
 & 0.838 & 0.400
 & 0.762 & \textbf{0.436}
 & 0.944 & \textbf{0.047}
 & 0.935 & 0.364 \\

 & \deepsight{}
 & \underline{0.808} & 0.221
 & 0.747 & 0.386
 & 0.942 & \cellcolor{red!15}0.560
 & 0.867 & \cellcolor{red!15}0.774 \\

 & \balance{}
 & 0.840 & 0.490
 & 0.770 & \cellcolor{red!15}0.504
 & 0.941 & \cellcolor{red!15}0.457
 & 0.869 & 0.457 \\

 & \spp{}
 & 0.835 & \cellcolor{red!15}0.525
 & 0.749 & \cellcolor{red!15}0.689
 & 0.943 & 0.048
 & 0.931 & \textbf{0.047} \\

 & \mmad{}
 & 0.853 & \textbf{0.184}
 & 0.764 & \cellcolor{red!15}0.625
 & 0.943 & 0.067
 & 0.923 & \cellcolor{red!15}0.801 \\

 & \ubar{}
 & 0.813 & \cellcolor{red!15}0.799
 & \underline{0.724} & \cellcolor{red!15}0.819
 & \underline{0.940} & 0.048
 & \underline{0.825} & 0.261 \\

 & \dfldual{}
 & 0.834 & 0.385
 & 0.770 & 0.343
 & 0.942 & 0.051
 & 0.920 & \cellcolor{red!15}0.798 \\

\midrule

\multirow{13}{*}{\rotatebox{90}{\Large \texttt{IBA}}}

 & \normclip{}
 & 0.849 & \cellcolor{red!15}0.785
 & 0.792 & \cellcolor{red!15}0.840
 & 0.943 & \cellcolor{red!15}0.806
 & 0.936 & \cellcolor{red!15}0.746 \\
 
 & \krum{}
 & 0.836 & \cellcolor{red!15}0.775
 & 0.713 & \cellcolor{red!15}0.793
 & 0.940 & 0.046
 & 0.922 & \cellcolor{red!15}0.841 \\

 & \trim{}
 & 0.849 & \cellcolor{red!15}0.810
 & 0.788 & \cellcolor{red!15}0.870
 & 0.945 & \cellcolor{red!15}0.667
 & 0.941 & \cellcolor{red!15}0.664 \\

 & \scclip{}
 & 0.828 & \cellcolor{red!15}0.809
 & 0.747 & \cellcolor{red!15}0.810
 & 0.944 & \cellcolor{red!15}0.588
 & 0.927 & \cellcolor{red!15}0.866 \\

 & \weakdp{}
 & 0.842 & \cellcolor{red!15}0.800
 & 0.775 & \cellcolor{red!15}0.870
 & 0.943 & \cellcolor{red!15}0.844
 & 0.938 & \cellcolor{red!15}0.804 \\

 & \abalance{}
 & 0.833 & \cellcolor{red!15}0.904
 & 0.758 & \cellcolor{red!15}0.912
 & 0.935 & 0.082
 & 0.929 & \textbf{0.370} \\

 & \flame{}
 & 0.848 & \cellcolor{red!15}0.874
 & 0.762 & \cellcolor{red!15}0.882
 & 0.934 & 0.100
 & 0.937 & \cellcolor{red!15}0.679 \\

 & \deepsight{}
 & \underline{0.810} & \cellcolor{red!15}0.677
 & 0.744 & \cellcolor{red!15}0.782
 & 0.935 & 0.497
 & 0.895 & \cellcolor{red!15}0.554 \\

 & \balance{}
 & 0.835 & \cellcolor{red!15}0.727
 & 0.770 & \cellcolor{red!15}0.857
 & 0.940 & 0.465
 & 0.910 & \cellcolor{red!15}0.571 \\

 & \spp{}
 & 0.835 & \cellcolor{red!15}0.945
 & 0.739 & \cellcolor{red!15}0.938
 & 0.943 & 0.291
 & 0.933 & 0.397 \\

 & \mmad{}
 & 0.850 & \cellcolor{red!15}0.691
 & 0.770 & \cellcolor{red!15}0.769
 & 0.937 & 0.083
 & 0.929 & \cellcolor{red!15}0.833 \\

 & \ubar{}
 & 0.819 & \cellcolor{red!15}0.723
 & \underline{0.710} & \cellcolor{red!15}0.807
 & \underline{0.933} & 0.046
 & \underline{0.830} & 0.402 \\

 & \dfldual{}
 & 0.843 & \cellcolor{red!15}0.757
 & 0.760 & \cellcolor{red!15}0.817
 & 0.944 & \textbf{0.044}
 & 0.925 & \cellcolor{red!15}0.765 \\

\bottomrule
\end{tabular}}
\end{table*}
\endgroup

\noindent\textbf{RQ2. How do established stealthy BAs (\texttt{A3FL, IBA, Neurotoxin}) behave in DFL settings?} We evaluate these attacks against 13 DFL-compatible defenses on \texttt{CIFAR-10} and
\texttt{GTSRB}, under IID and non-IID client partitions. Table~\ref{tab:rq2-results} summarizes our findings using a [30,60] attack window.

\textit{Overall attack potency.}
Across defenses and datasets, attack strength consistently ranks
\textbf{IBA $>$ A3FL $>$ Neurotoxin}, with mean ASRs of
$0.57/0.33/0.22$ under IID and $0.74/0.52/0.35$ under non-IID,
respectively. \texttt{IBA} causes an ASR $\geq 0.500$ failure in 75\% of
defense--dataset--partition combinations, versus 35\% for \texttt{A3FL} and
21\% for \texttt{Neurotoxin}. \texttt{IBA} and \texttt{A3FL} are more effective because their
adaptive, optimized triggers are designed to
evade filtering, whereas \texttt{Neurotoxin} relies on a
static, low-magnitude gradient mask. Static
BAs (BadNet, DBA, and Scaling) are generally less effective
in our setting and are omitted for brevity.

\texttt{IBA} \textit{breaks every defense on} \texttt{CIFAR-10.}
Against \texttt{IBA}, all defenses exceed $0.677$ ASR on
\texttt{CIFAR-10} in \textit{both} partitions -- including
\abalance{} and \flame{}, which are
otherwise the strongest defenses overall. This is not primarily a
heterogeneity effect: the same defenses that fail outright on
\texttt{CIFAR-10} IID recover on \texttt{GTSRB} IID, where
\krum{}, \abalance{}, \flame{}, \mmad{}, \ubar{}, and \dfldual{} all
hold \texttt{IBA} below $0.10$ ASR. The gap points to a dataset-level
vulnerability -- \texttt{CIFAR-10}'s lower per-class margin
under a 10-way task -- that \texttt{IBA}'s optimized trigger exploits
independently of the aggregation rule. Under non-IID \texttt{GTSRB},
this protection narrows further: only \abalance{} and
\ubar{} remain below the failure threshold; every
other defense, gives way.

\textit{Defense effectiveness.}
Averaged over all attacks, datasets, and partitions, \abalance{} and
\flame{} remain the most consistent defenses (mean ASR $0.347$ and
$0.341$), but this is a story of consistency rather than IID
dominance: under IID data alone, \dfldual{} and \krum{}
 are actually marginally lower than \abalance{}/\flame{}, aided by strong \texttt{Neurotoxin/A3FL} suppression on
\texttt{GTSRB}. That IID advantage evaporates under non-IID data,
where \krum{} and \mmad{} post the largest degradations of any
defense, followed by \dfldual{}, consistent with their reliance on inter-client
similarity or tightly clustered updates -- an assumption that breaks
down once benign updates diverge naturally. \scclip{} is the weakest
defense overall (mean ASR $0.685$), confirming that self norm
clipping alone is insufficient against adaptive triggers.
\deepsight{}, \balance{}, and \spp{} sit in a middle tier, with
smaller IID-to-non-IID increases but higher absolute non-IID ASR
than \abalance{}/\flame{}.

\textit{Accuracy--robustness trade-off.}
Aggressive filtering does not consistently buy robustness.
\deepsight{} has the worst Min.\,ACC on \texttt{CIFAR-10} IID across
all three attacks ($0.810$) while its mean ASR ($0.450$) is only
mid-tier, the weakest trade-off of any defense. \ubar{} shows the
opposite pattern of \emph{inconsistency}: it posts the worst Min.\,ACC
under non-IID.

\begin{tcolorbox}[breakable, boxrule=0pt, colframe=gray!40]
\textbf{Takeaways.}
\begin{inparaenum}[(i)]
\textit{\item \texttt{IBA} defeats every evaluated defense on \texttt{CIFAR-10}}, regardless of
aggregation rule or client partition (min.\ ASR $0.68$ across all
defenses) -- a dataset-level vulnerability, not merely a non-IID artifact.
\textit{\item} \abalance{} \textit{and} \flame{} \textit{are the most
consistent defenses on average}, but their IID-only edge over \dfldual{} and
\krum{} is small and disappears once \textit{IBA}-on-\textit{CIFAR-10} is included.
\textit{\item Similarity- and clustering-based defenses are non-IID brittle}:
\krum{}, \mmad{}, and \dfldual{} show the largest IID-to-non-IID ASR
increases.
\end{inparaenum}
\end{tcolorbox}

\begin{figure*}[t]
    \centering
    \includegraphics[width=0.9\textwidth]{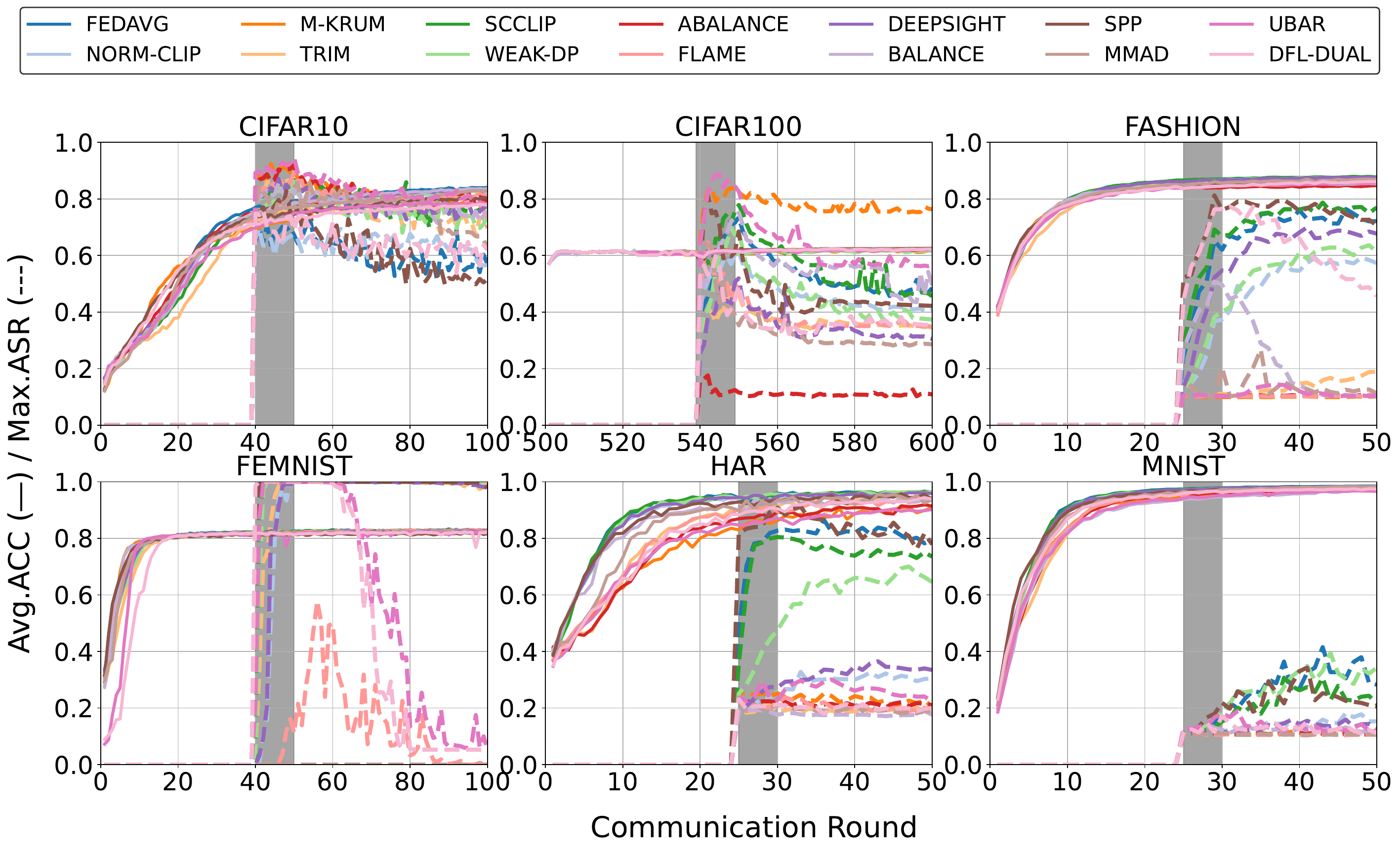}
    \caption{Impact of datasets and models under attacks.}
    \label{fig:datasets}
\end{figure*}

\noindent\textbf{\textbf{RQ3.} What is the impact of data and model characteristics on robustness in DFL?} We study the effectiveness of all the defenses across multiple datasets, pairing each with a distinct attack (\texttt{IBA} on \texttt{CIFAR-10}/\texttt{CIFAR-100}, \texttt{A3FL} on \texttt{FEMNIST}/\texttt{MNIST}, and \texttt{Neurotoxin} on \texttt{HAR}/\texttt{FASHION}) to probe how data and model properties, independent of the attack itself, shape robustness. Figure~\ref{fig:datasets} visualizes the impact of each dataset under short attack windows. We find that dataset characteristics fundamentally shape backdoor dynamics.

\textit{First,} simple, low-dimensional datasets attacked with \texttt{A3FL} on \texttt{MNIST} exhibit limited backdoor persistence: nearly every defense drives durability to 0 and final ASR below 0.22, with only \weakdp{} failing to suppress the attack. In contrast, \texttt{FEMNIST} under the same attack shows a sharp split: effective defenses (\krum{}, \abalance{}, \balance{}, \mmad{}) fully block the backdoor from round one (durability~=~0), while clipping- and similarity-based defenses fail completely, leaving ASR pinned near 1.00 for the full 49-round window; \flame{}, \ubar{}, and \dfldual{} fall in between, eventually suppressing the attack only after a long delay. This shows that dataset and task characteristics, beyond input dimensionality alone, governs how long a backdoor can persist once it evades early detection. On \texttt{CIFAR-100}, attacked with \texttt{IBA}, most defenses lower Max.ASR relative to \texttt{CIFAR-10} yet durability remains high (>40) for nearly all of them. \textit{Second,} \texttt{HAR} sequences under \texttt{Neurotoxin} reveal a distinct vulnerability rooted in the temporal alignment of benign gradients: this alignment enables low-norm, temporally consistent poisoning that bypasses \scclip{} (Final.ASR~=~0.74), \weakdp{} (0.65), and \deepsight{} (0.33), all of which retain full durability, whereas \trim{}, \balance{}, \mmad{}, and \dfldual{} suppress the same attack down to final ASR~$\leq$~0.20. At the same time, the high false-positive rate of \ubar{} degrades sharply Min.ACC (to 0.76) while still failing to fully contain the attack.

\begin{tcolorbox}[breakable, boxrule=0pt, colframe=gray!40]
\textbf{Takeaways.}
\begin{inparaenum}[(i)]
    \textit{\item Dataset properties fundamentally shape backdoor dynamics}: on \texttt{CIFAR-100} (\texttt{IBA}), defenses can reduce Max.ASR while durability stays high, giving a misleading impression of robustness.
    \textit{\item Attack-dataset pairing matters as much as the defense}: \texttt{A3FL} is fully contained by robust aggregation on \texttt{MNIST}/\texttt{FEMNIST}, while \texttt{IBA} evades every defense on \texttt{CIFAR-10}.
    \textit{\item Temporal structure acts as a double-edged system factor}: on \texttt{HAR}, alignment of benign gradients aids convergence but also enables \texttt{Neurotoxin} to reach high, durable ASR.
    \textit{\item The cost of aggressive filtering} in terms of Min.ACC degradation varies substantially across datasets.
\end{inparaenum}
\end{tcolorbox}

\begin{figure*}[th]
    \centering
    \resizebox{\linewidth}{!}{
    \subfloat[\scriptsize  Regular-(20,7)]{
        \includegraphics[scale=0.13]{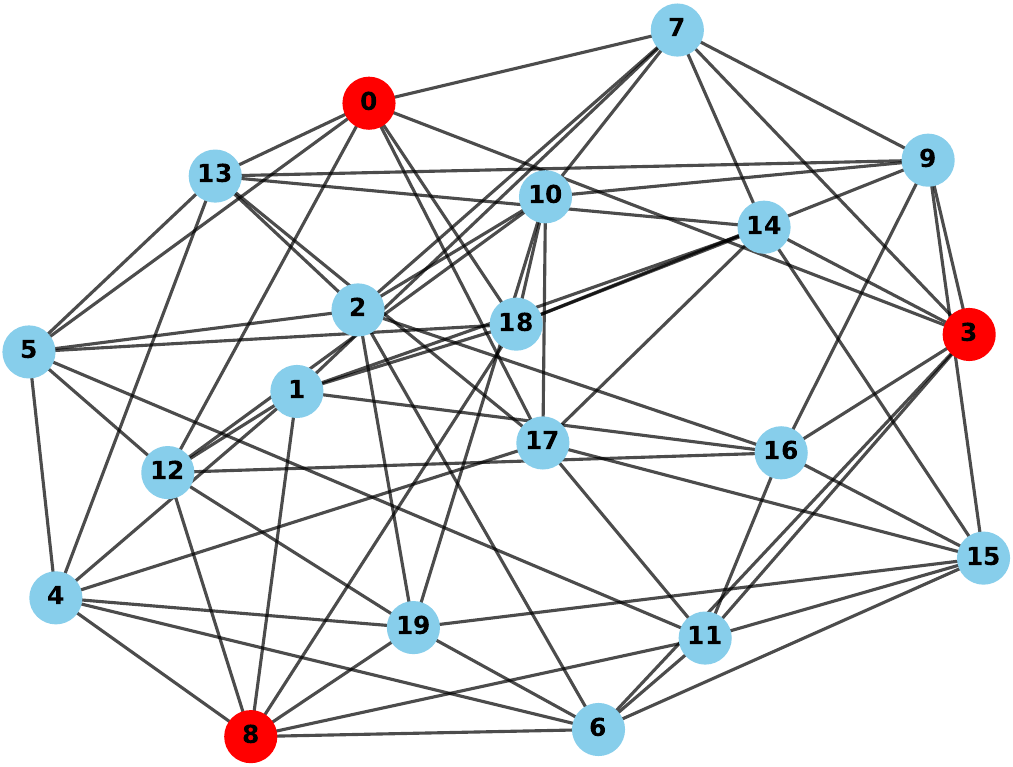}
        \label{fig:graph_regular}
    }
    \subfloat[\scriptsize Regular-(50,10)]{
        \includegraphics[scale=0.13]{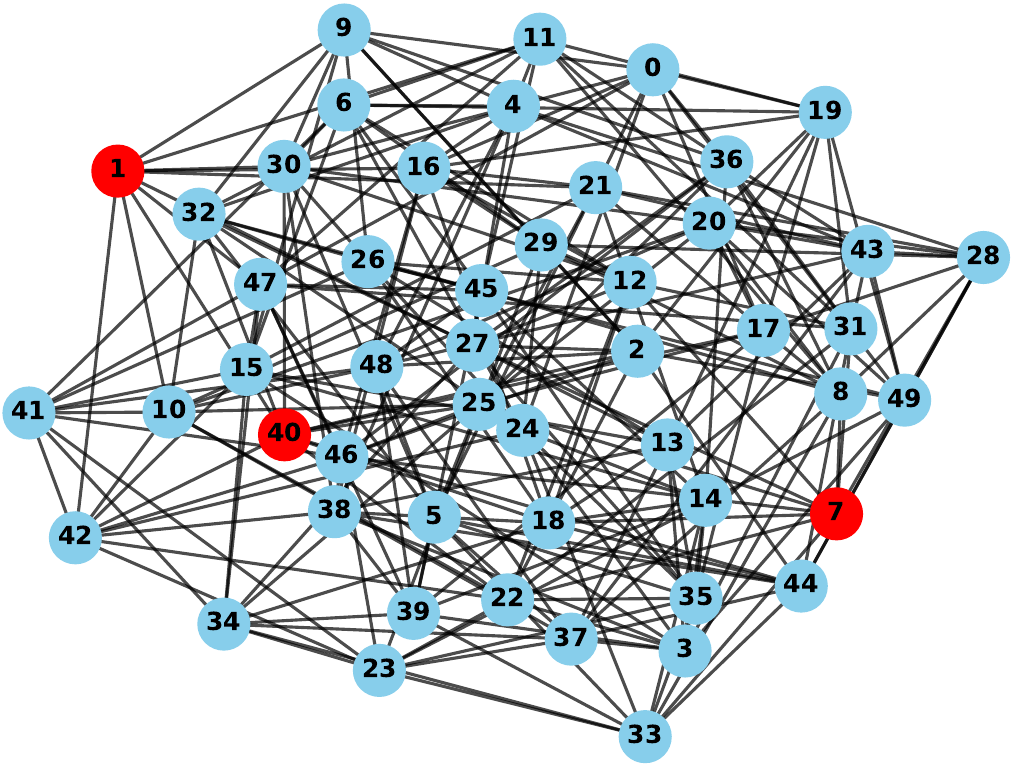}
        \label{fig:graph_regular_50}
    }
    \subfloat[\scriptsize  Ring-(20,7)]{
        \includegraphics[scale=0.13]{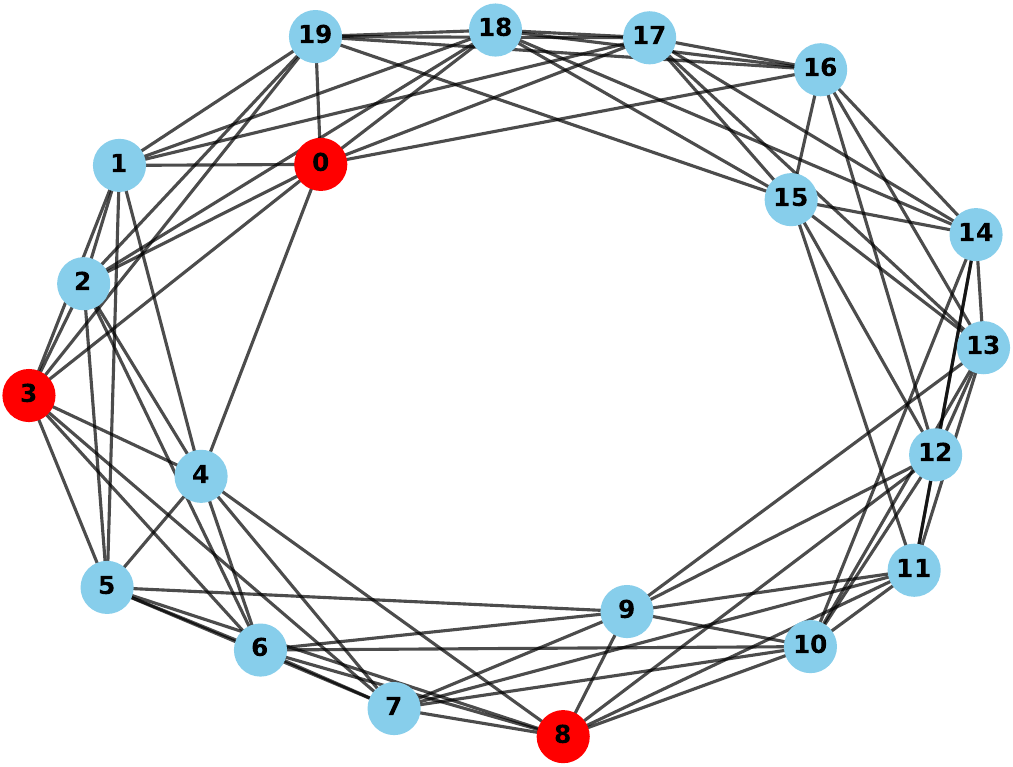}
        \label{fig:graph_ring}
    }
    \subfloat[\scriptsize Watts-(20,7)]{
        \includegraphics[scale=0.13]{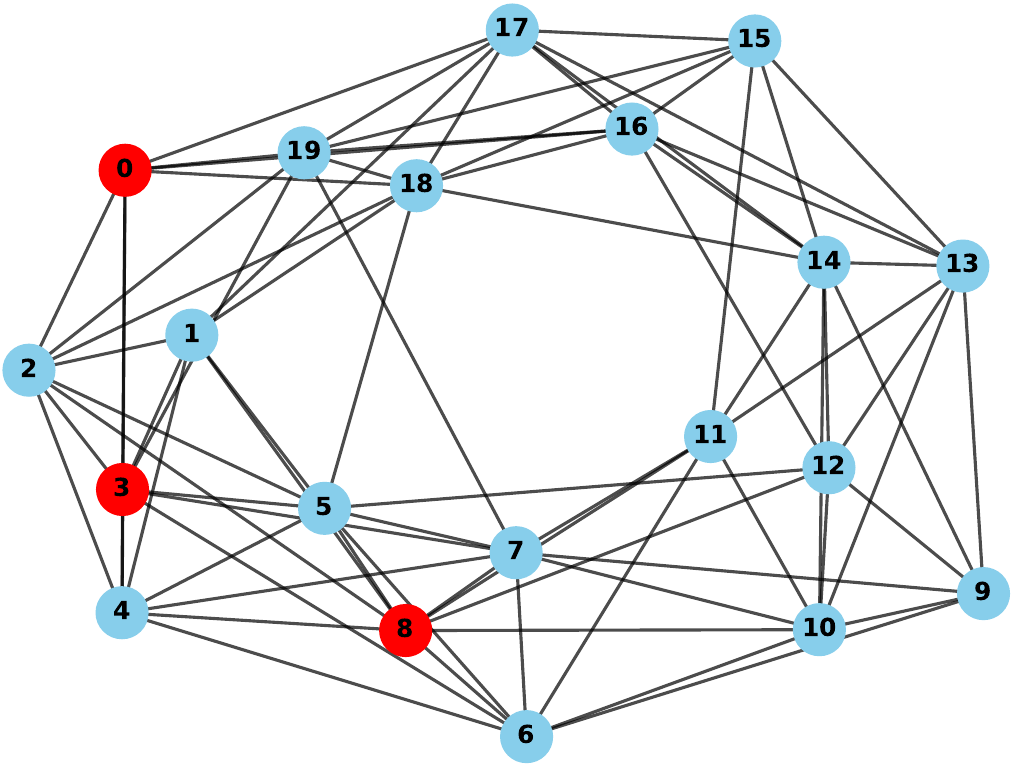}
        \label{fig:graph_watts}
    }
    \subfloat[\scriptsize  Barabasi-(20,7)]{
        \includegraphics[scale=0.13]{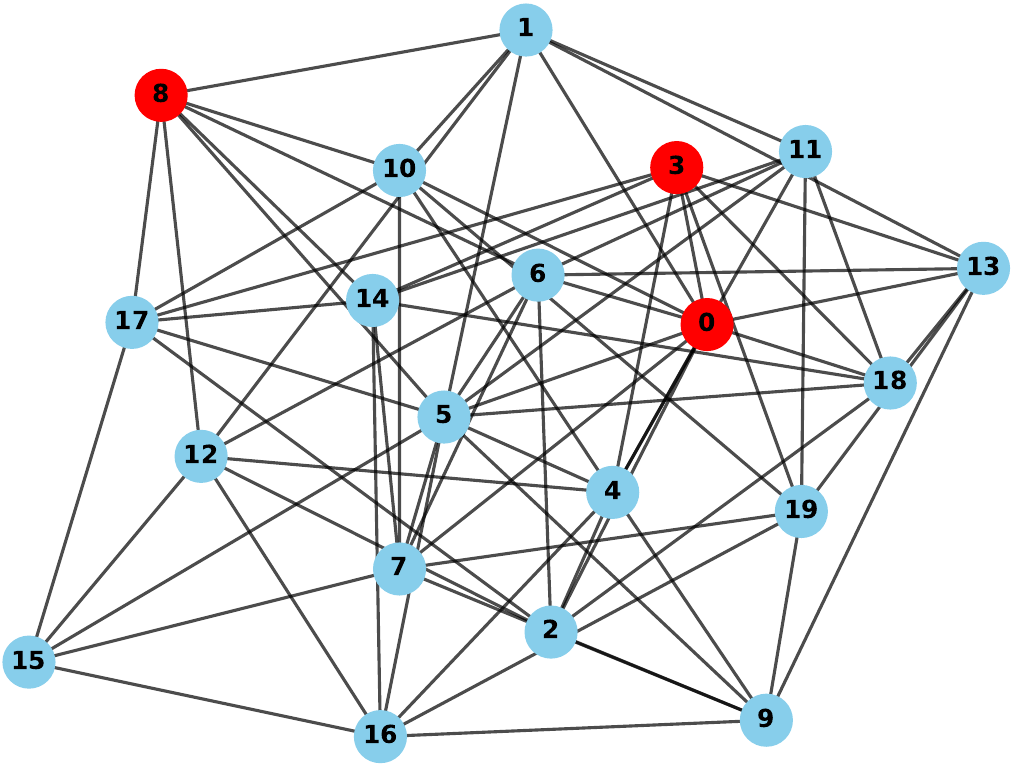}
        \label{fig:graph_barabasi}
    }
    \subfloat[\scriptsize Erdős-(20,7)]{
        \includegraphics[scale=0.13]{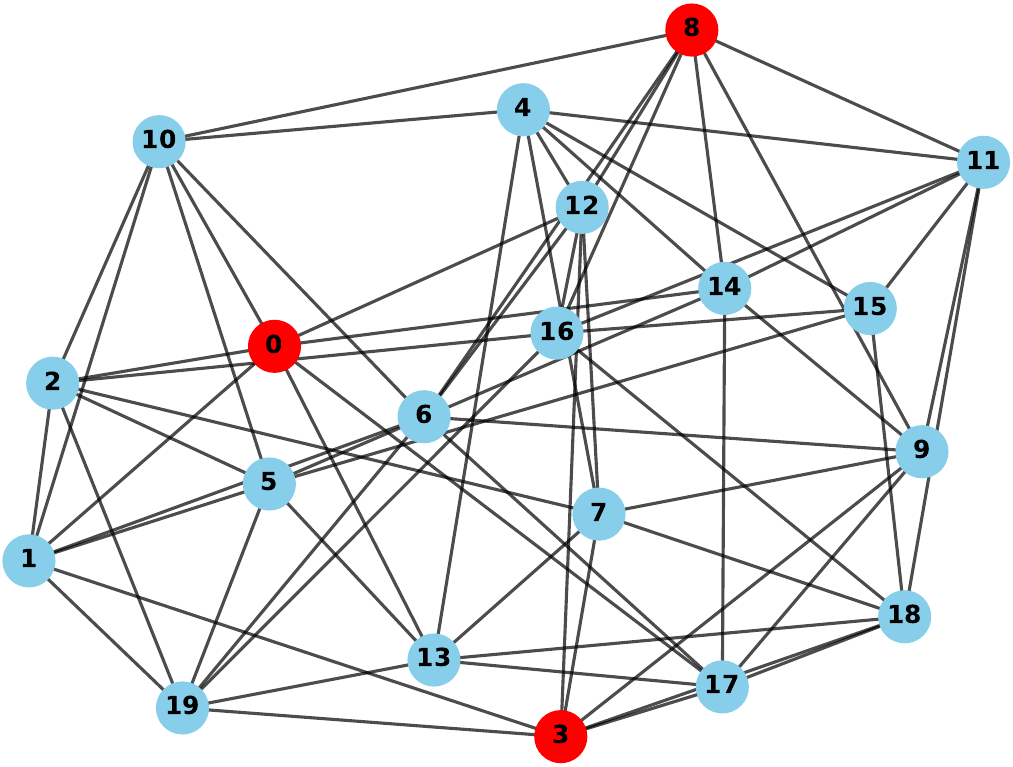}
        \label{fig:graph_erdos}
    }}
    \caption{Different communication graphs.}
    \label{fig:graphs}
\end{figure*}

\begin{figure*}[th]
    \centering
    \includegraphics[scale=0.25]{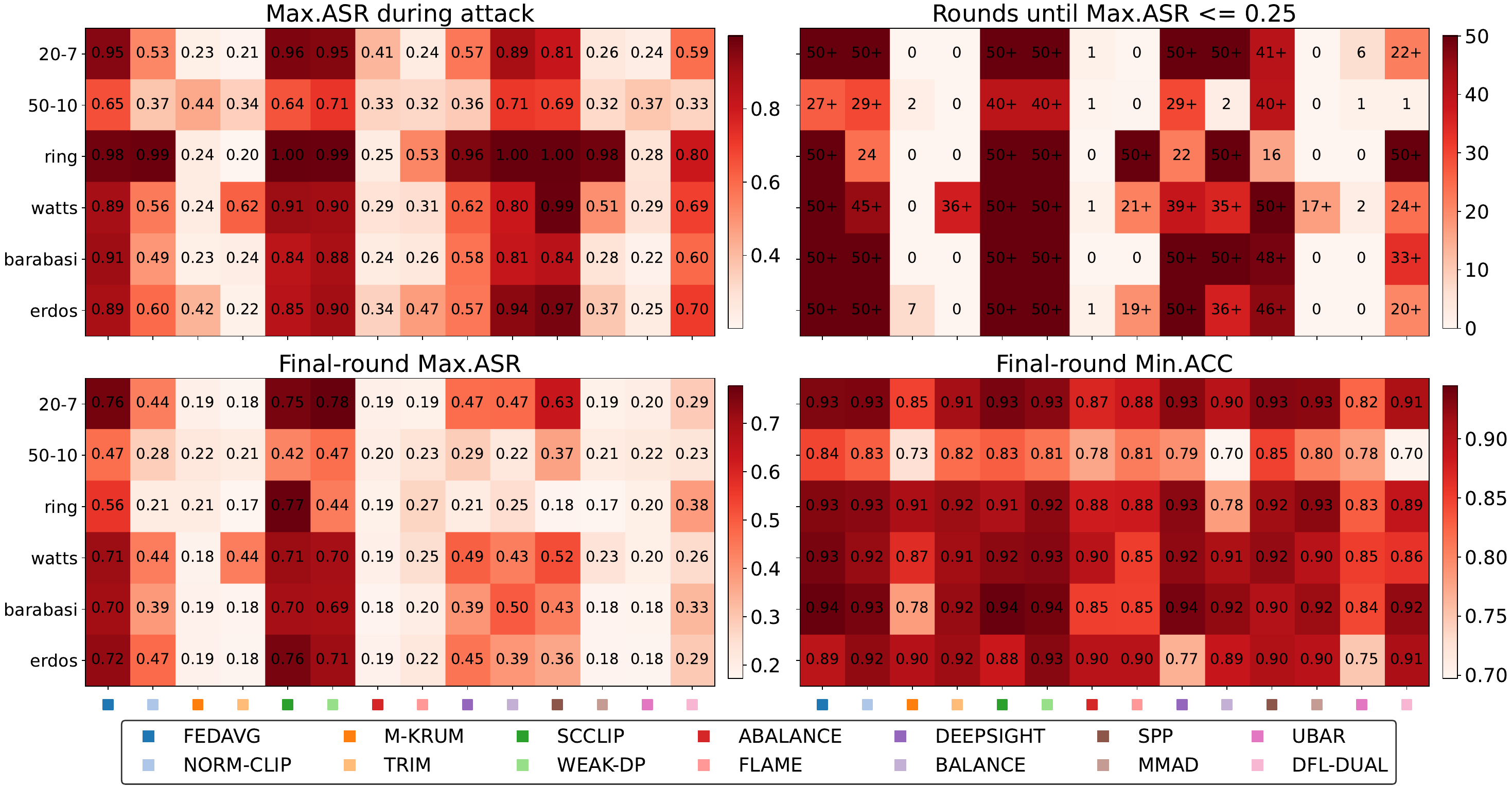}
    \caption{Impact of the communication graph.}
    \label{fig:impact_graph}
\end{figure*}

\noindent\textbf{\textbf{RQ4.} What is the impact of the DFL graph topology?}
We evaluate all defenses across different graph topologies (Figure~\ref{fig:graphs})
using 3 different seeds (21, 42, 84) to assess how the network structure influences
the robustness of each defense. The heat maps in Figure~\ref{fig:impact_graph} report the results of \texttt{HAR} under \texttt{Neurotoxin} (attack
window = [25,50]) to show the speed at which the global model recovers after
contamination.  Without any defense, attack severity already depends strongly on topology: peak ASR ranges from 65.3\% on \texttt{regular-(50-10)} (large, structured graph) up to 97.8\% on \texttt{ring} (sparse graph), with \texttt{barabasi},
\texttt{erdos}, and \texttt{watts} clustering around 88--95\%. Defense
generalization is highly topology-dependent even on this easiest setting:
\abalance{} and \ubar{} are the only defenses that suppress the
backdoor (in 0--6 rounds, final ASR $\leq$ 21\%) on all six topologies, while
distance-based defenses such as \trim{} and \krum{} succeed almost
everywhere but fail on specific graphs, \eg, \trim{} suppresses instantly
(0 rounds) on \texttt{barabasi}, \texttt{erdos}, \texttt{regular-(20-7)},
\texttt{regular-(50-10)}, and \texttt{ring}, yet never recovers within the training
horizon on \texttt{watts} (36.3 rounds, censored, final ASR 43.9\%). Conversely,
CFL defenses (\normclip{}, \deepsight{}, \weakdp{}, \spp{}) and the DFL-specific \balance{}/\dfldual{} remain unsuppressed (50+ rounds, censored) on the shortcut-rich topologies (\texttt{barabasi}, \texttt{erdos}, \texttt{regular-(20-7)}, \texttt{watts}) but recover on the sparse \texttt{ring} topology (\eg, \normclip{}: 24 rounds,
\deepsight{}: 22 rounds, \spp{}: 16 rounds), where the same defenses
fail elsewhere.


\begin{tcolorbox}[breakable, boxrule=0pt,
  colframe=gray!40]
\textbf{Takeaways.}
\textit{(i) Graph topology governs both attack amplification and recovery}:
structured graphs can either localize an attack (\texttt{regular-(50-10)}, 65.3\%
peak ASR) or prolong it (\texttt{ring}, 97.8\% peak ASR), while shortcut-rich
topologies (\texttt{barabasi}, \texttt{erdos}, \texttt{watts}) both accelerate
diffusion (peak ASR $\geq$ 88\%) and
mitigation (for effective defenses). \textit{(ii) Defenses developed for both CFL and
DFL fail to generalize across graph topologies}: only \abalance{} and
\ubar{} suppress the backdoor on every topology tested. \textit{(iii) These trends reflect a favorable setting}: \texttt{HAR} is one of the lightest tasks, paired with \texttt{Neurotoxin}, one of the least stealthy attacks. Under stealthier attacks (\texttt{IBA} in \textbf{RQ2}) and on more challenging tasks (\texttt{GTSRB} and \texttt{CIFAR-10}), all defenses fail regardless of topology, meaning robustness gains narrow as attack stealth and task difficulty increase.
\end{tcolorbox}

\section{Conclusion}
\label{sec:conclusion}

In this paper, we presented BackDFL, a unified benchmark of backdoor attacks and defenses in decentralized federated learning (DFL) to adaptive backdoor attacks. Using the proposed BackDFL reproducible suite, we show that robust peer-to-peer DFL is highly susceptible to backdoor contamination due to unrestricted diffusion of malicious updates. State-of-the-art Byzantine-robust DFL methods fail even under modest malicious rates (15\%) and realistic non-IID distributions. We further showed that adapting a lightweight defense such as BALANCE, by making its acceptance criterion responsive to the actual neighbor updates, can improve resilience while avoiding overly aggressive filtering, providing a practical, low-computation alternative to more costly centralized defenses. Finally, our analysis reveals that defenses developed for both FL and DFL fail to generalize across different graph topologies, highlighting the need for adaptive mechanisms that can maintain robustness under diverse networks.




\bibliographystyle{splncs04}
\bibliography{refs}

\appendix
\section{Detailed Experimental Settings}
\label{app:hyper}

The models listed in Table~\ref{tab:configurations} are standard models adapted to each dataset's characteristics. 
For \texttt{MNIST}, we use a SimpleCNN composed of two 5$\times$5 convolutional layers (6 and 16 filters), followed by ReLU activations, 2$\times$2 average pooling, and fully connected layers of sizes 120, 84, and 10. For \texttt{FashionMNIST}, we employ a lightweight convolutional neural network (FashionCNN) consisting of two convolutional blocks with 32 and 64 filters ($3\times3$), each followed by ReLU activation and $2\times2$ max-pooling, and two fully connected layers (128 units and a 10-class output), with dropout (0.5) applied before the final classification layer.
For \texttt{FEMNIST}, we employ a LeNet-5 consisting of two convolutional layers with 6 and 16 filters (5$\times$5), each followed by average pooling, and three fully connected layers with 120 and 84 hidden units before the final classification layer. For \texttt{CIFAR-10}, we adopt a ResNet-18 variant where all BatchNorm layers are replaced with GroupNorm; the first convolution is a 3$\times$3 stride-1 layer without an initial max-pooling stage, followed by a 10-class classification head. For \texttt{GTSRB}, the GTSRB-CNN consists of two convolutional blocks: the first with 32 filters (5$\times$5, ReLU, max-pooling, dropout 0.25), and the second with 64 filters (3$\times$3, ReLU, max-pooling, dropout 0.25), followed by a fully connected layer of size 512 with ReLU, dropout 0.5, and a 43-class output layer. For \texttt{HAR}, we employ a lightweight multilayer perceptron (HAR-MLP) composed of two hidden fully connected layers with 256 and 128 units, respectively, each followed by ReLU activations, and a final 6-class output layer, taking the 561-dimensional feature vector as input.   

\begin{table*}[t]
\centering
\small
\caption{Dataset, model, and training hyperparameter configurations.}
\label{tab:configurations}
\resizebox{0.65\textwidth}{!}{
\begin{tabular}{l l c c c c c c c c}
\toprule
\textbf{Dataset} & \textbf{Model} & \textbf{\#Classes} & \textbf{Input} &
\textbf{Optimizer} & \boldmath$\eta$ & \boldmath$B$ & \boldmath$E$ & \textbf{Rounds} \\
\midrule
\texttt{MNIST}         & SimpleCNN      & 10  & $1\times28\times28$ & SGD & 0.01 & 32 & 1  & 50 \\
\texttt{Fashion}  & FashionCNN   & 10  & $1\times28\times28$ & SGD & 0.01 & 32 & 5  & 50 \\
\texttt{FEMNIST}       & LeNet-5        & 62  & $1\times28\times28$ & SGD & 0.01 & 32 & 10 & 100 \\
\texttt{CIFAR-10}      & ResNet18-GN   & 10  & $3\times32\times32$ & SGD & 0.01 & 32 & 5  & 100 \\
\texttt{CIFAR-100}     & ResNet18-GN   & 100 & $3\times32\times32$ & SGD & 0.01 & 32 & 5  & 600$^*$ \\
\texttt{GTSRB}         & GTSRB-CNN     & 43  & $3\times32\times32$ & SGD & 0.01 & 32 & 5  & 50--100 \\
\texttt{HAR}           & HAR-MLP       & 6   & 561                 & SGD & 0.01 & 32 & 5  & 50--300 \\
\bottomrule
\end{tabular}}

\scriptsize $^*$ 500 benign rounds, followed by 100 rounds from the benign checkpoint with a scheduled attack. 
\end{table*}

\section{Sensitivity Analysis}
\label{app:factors}

\begin{figure}[t]
    \centering
    \includegraphics[width=0.9\columnwidth]{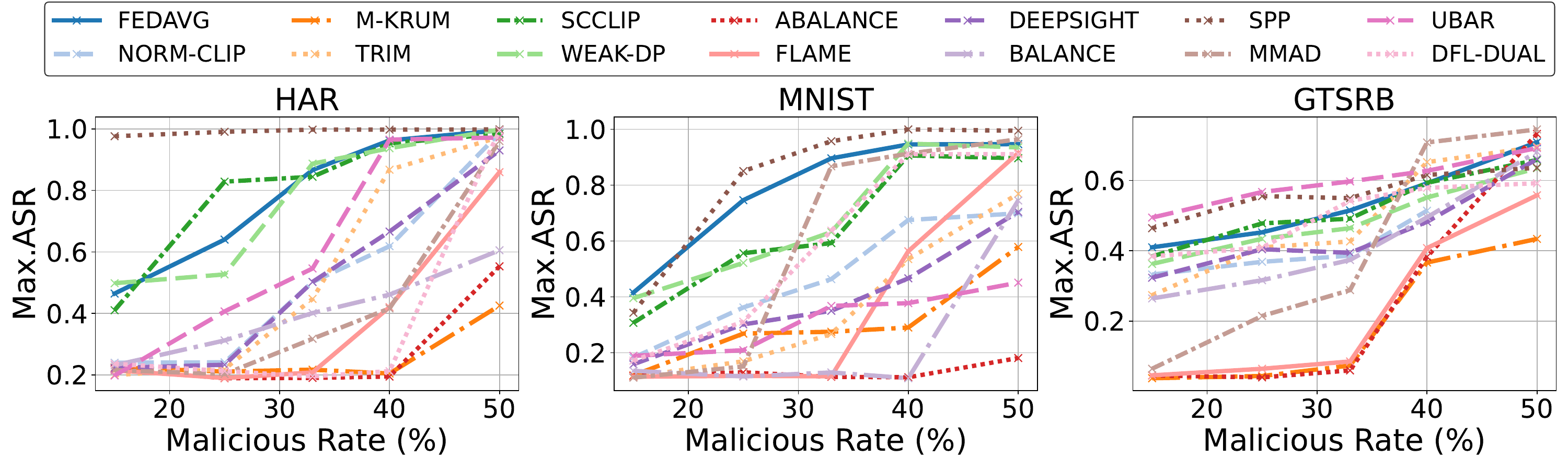}
    \caption{Impact of the malicious rate.}
    \label{fig:rateimpact}
\end{figure}
\noindent\textbf{Impact of malicious rate.} We evaluate all defenses under increasing global malicious ratios $\beta_M$ (15--50\%) and a short attack window. Figure~\ref{fig:rateimpact} shows results on three light tasks ((\texttt{A3FL} on \texttt{MNIST}/\texttt{GTSRB}, \texttt{Neurotoxin} on \texttt{HAR})). Robust methods such as \krum{}, \abalance{}, and \flame{} maintain low ASR at 15--25\% malicious clients; on \texttt{MNIST}, their Final.ASR remains at 0.11--0.12. At 40\%, they still achieve a Final.ASR of 0.20 on \texttt{HAR}, but degrade sharply beyond 33\% on \texttt{GTSRB}. Overall, \abalance{} is \textit{the most stable defense} across datasets and malicious rates.

\begin{figure}[t]
    \centering
    \includegraphics[width=\columnwidth]{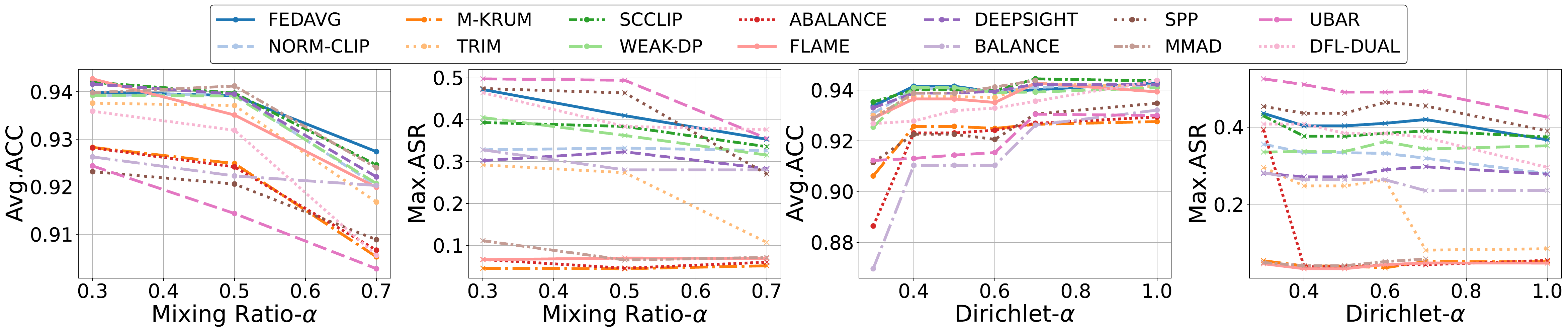}
    \caption{Impact of mixing ratio (\textbf{Left}) and non-IID degree (\textbf{Right}).}
    \label{fig:mix_alpha}
\end{figure}

\noindent\textbf{Impact of the mixing ratio.} We evaluate all defenses under increasing mixing ratios $\alpha$ in Eq.~\eqref{eq:dfl-update} to study the effect of client reliance on neighbors versus self-training. Figure~\ref{fig:mix_alpha} (\textbf{Left}) reports the results on \texttt{GTSRB} under a short \texttt{A3FL} attack window of 6 rounds . As $\alpha$ decreases (i.e., stronger reliance on neighbor aggregation), backdoor persistence grows for permissive defenses. For example, at $\alpha = 0.3$, \scclip{}, \normclip{}, \weakdp{}, and \ubar{} reach Max.ASR values of 0.39–0.50. In contrast, strong defenses such as \flame{} and \abalance{} maintain strong robustness (Max.ASR $\leq 0.07$). However, like all methods, stronger self reliance decreases their utility. 

\noindent\textbf{Impact of data heterogeneity.} We evaluate the robustness of all defenses under decreasing heterogeneity $Dirichlet$-$\alpha$ (0.3–1.0) on \texttt{GTSRB} with only $15\%$ \texttt{A3FL} malicious rate. Figure~\ref{fig:mix_alpha} (\textbf{Right}) reports Max.ASR and Avg.ACC for all methods. Robust defenses such as \flame{} and \mmad{} maintain low Max.ASR across heterogeneity levels (0.03–0.06) while preserving high Avg.ACC (0.92–0.95). Similarly, \abalance{} remains highly robust, with Max.ASR $\leq$ 0.05 and minimal accuracy degradation. In contrast, smoothing-based defenses (\normclip{}, \scclip{}, \weakdp{}) deteriorate markedly under high data heterogeneity. \spp{} and \ubar{} perform even worse, attaining Max.ASR in the range of 0.36–0.46 with strong persistence. Although \ubar{} uses a second-stage filtering based on local loss signals, this provides little benefit under targeted BAs, where \textit{loss separation becomes ineffective}.

\begin{table*}[t]
\begin{minipage}{0.54\textwidth}
\caption{\small Impact of local epochs.}
\centering
\Large
\label{tab:epochs_a3fl_neurotoxin}
\resizebox{0.95\columnwidth}{!}{
\begin{tabular}{llcccccc}
\toprule
& & \multicolumn{3}{c}{\texttt{Min.ACC}} & \multicolumn{3}{c}{\texttt{Max.ASR}} \\
\cmidrule(lr){3-5} \cmidrule(lr){6-8}
 \textbf{Dataset} & \textbf{Defense} & 1 & 5 & 10 & 1 & 5 & 10  \\
\midrule
\multirow{3}{*}{\texttt{GTSRB}}
& \none{}
& 0.921 & 0.947 & 0.943
& \cellcolor{red!15}0.765 & \cellcolor{red!15}0.584
& \cellcolor{red!15}0.464 \\
& \flame{}
& \underline{0.919} & 0.935 & \underline{0.933}
& \textbf{0.102} & 0.364
& 0.478 \\
& \abalance{}
& 0.923 & \underline{0.932} & 0.935
& 0.127 & \textbf{0.074} & \textbf{0.389} \\
\midrule
\multirow{3}{*}{\texttt{HAR}}
& \none{}
& 0.974 & 0.971 & 0.970
& \cellcolor{red!15}0.999 & \cellcolor{red!15}0.981
& \cellcolor{red!15}0.957 \\
& \flame{}
& 0.962 & 0.960 & 0.954
& \textbf{0.218} & 0.239 & \textbf{0.228} \\
& \abalance{}
& \underline{0.954} & \underline{0.949} & \underline{0.944}
& 0.296 & \textbf{0.235} & 0.231 \\
\bottomrule
\end{tabular}}
\end{minipage}
\begin{minipage}{0.43\textwidth}
\caption{\small Impact of batch size.}
\centering
\Large
\label{tab:batch_size_a3fl_neurotoxin}
\resizebox{0.94\columnwidth}{!}{
\begin{tabular}{ccccccccc}
\toprule
 \multicolumn{4}{c}{\texttt{Min.ACC}} & \multicolumn{4}{c}{\texttt{Max.ASR}} \\
\cmidrule(lr){1-4} \cmidrule(lr){5-8}
   16 & 32 & 48 & 64 & 16 & 32 & 48 & 64  \\
\midrule

 0.951 & 0.947 & 0.938 & 0.929
& \cellcolor{red!15}0.612 & \cellcolor{red!15}0.584
& \cellcolor{red!15}0.536 & \cellcolor{red!15}0.510 \\

  \underline{0.946} & 0.935 &  \underline{0.934} & 0.926
& \cellcolor{red!15}\textbf{0.513} & 0.364
& 0.323 & 0.104 \\

 0.950 &  \underline{0.932} & 0.938 & \underline{0.920}
& \cellcolor{red!15}0.576 & \textbf{0.074}
& \textbf{0.229} & \textbf{0.103} \\
\midrule

 0.972 & 0.971 & 0.972 & 0.971
& \cellcolor{red!15}0.896 & \cellcolor{red!15}0.981
& \cellcolor{red!15}0.980 & \cellcolor{red!15}0.939 \\

 0.962 & 0.960 & 0.959 & 0.949
& \textbf{0.249} & 0.235
& 0.251 & \textbf{0.290} \\

 \underline{0.949} & \underline{0.949} & \underline{0.943} & \underline{0.934}
& 0.284 & \textbf{0.232} & \textbf{0.247} & 0.307 \\
\bottomrule
\end{tabular}}
\end{minipage}

\end{table*}

\noindent \textbf{Impact of local epochs.} We vary the number of local training epochs $(1,5,10)$ (Table~\ref{tab:epochs_a3fl_neurotoxin}) on \texttt{GTSRB} and \texttt{HAR}. Min.ACC is largely stable across all defenses and epoch settings, with only small changes as epochs increase/decrease. Thus, main-task utility is not heavily affected by local training length. Max.ASR, however, is far more sensitive: on \texttt{GTSRB}, \abalance{} shows more graceful degradation as local epochs increase, particularly on \texttt{HAR} , whereas \flame's robustness deteriorates more noticeably with longer local training. Increasing the number of local training epochs benefits
\abalance{} more than \flame{} as longer local training amplifies the magnitude and heterogeneity of benign updates, particularly under non-IID data, making a less adaptive acceptance criterion more difficult to calibrate. In contrast, \abalance{} adapts its threshold to the empirical distribution of received deviations while retaining the previous threshold as a temporal constraint.

\noindent \textbf{Impact of batch size.} We vary the local batch size $(16,32,48,64)$ (Table~\ref{tab:batch_size_a3fl_neurotoxin}) on \texttt{GTSRB} and \texttt{HAR} to assess robustness under different training batchs. Across both datasets, Min.ACC degrades only marginally as batch size increases (\eg, \texttt{GTSRB} \none{} drops from 0.951 to 0.929; \abalance{} from 0.950 to 0.920). In contrast, \texttt{Max.ASR} shows much larger fluctuations, especially without defense. Both \flame{} and \abalance{} suppress ASR across large batch sizes, with \abalance{} achieving the lowest ASR in most \texttt{GTSRB} settings while maintaining competitive accuracy. On HAR, \flame{} attains the overall lowest ASR values, but \abalance{} remains close behind and consistently more stable in accuracy. 

\end{document}